\documentclass[journal,twoside,web]{ieeecolor}
\usepackage{generic}
\usepackage{algorithmic}
\usepackage{textcomp}
\usepackage{graphicx, cite, amsmath, booktabs, multirow, adjustbox}
\usepackage{subfig, threeparttable, wrapfig}
\usepackage{amsmath, amsfonts, amssymb, pifont}
\usepackage[table]{xcolor}
\usepackage{colortbl}
\usepackage{arydshln}
\usepackage{diagbox}
\usepackage{stfloats}
\usepackage{bm}
\usepackage{makecell}
\usepackage[
    colorlinks,
    linkcolor=red,
    anchorcolor=blue,
    citecolor=green,
    bookmarks=false,
    ]{hyperref}
\usepackage{url}

\graphicspath{{imgs/}}
\def\BibTeX{{\rm B\kern-.05em{\sc i\kern-.025em b}\kern-.08em
    T\kern-.1667em\lower.7ex\hbox{E}\kern-.125emX}}
\newcommand{\bstscr}[1]{\textcolor{red}{\textbf{#1}}}
\newcommand{\spacedashline}{%
    \addlinespace[0.27em]
    \hdashline
    \addlinespace[0.27em]
}

\definecolor{mpblue}{RGB}{31,119,180}
\definecolor{mporange}{RGB}{255,127,14}
\definecolor{mpgreen}{RGB}{44,160,44}
\definecolor{mpred}{RGB}{214,39,40}
\definecolor{mppurple}{RGB}{148,103,189}
\definecolor{mpbrown}{RGB}{140,86,75}
\definecolor{mppink}{RGB}{227,119,194}
\definecolor{mpgray}{RGB}{127,127,127}
\definecolor{mpolive}{RGB}{188,189,34}
\definecolor{mpcyan}{RGB}{23,190,207}

\definecolor{mplightblue}{RGB}{174,199,232}
\definecolor{mplightorange}{RGB}{255,187,120}
\definecolor{mplightgreen}{RGB}{152,223,138}
\definecolor{mplightred}{RGB}{255,152,150}
\definecolor{mplightpurple}{RGB}{197,176,213}
\definecolor{mplightbrown}{RGB}{196,156,148}
\definecolor{mplightpink}{RGB}{247,182,210}
\definecolor{mplightgray}{RGB}{199,199,199}
\definecolor{mplightolive}{RGB}{219,219,141}
\definecolor{mplightcyan}{RGB}{158,218,229}

\IEEEaftertitletext{\vspace{-1.5\baselineskip}}

\begin{document}
\title{
    FISHER: A Foundation Model for Multi-Modal Industrial Signal Comprehensive Representation
}
\author{Pingyi Fan,~\IEEEmembership{Senior Member,~IEEE,} Anbai Jiang, Shuwei Zhang, Xinhu Zheng, Zhiqiang Lv,~\IEEEmembership{Member,~IEEE,} Bing Han, Wenrui Liang,~\IEEEmembership{Student Member,~IEEE,} Junjie Li, Wei-Qiang Zhang,~\IEEEmembership{Senior Member,~IEEE,} Yanmin Qian,~\IEEEmembership{Senior Member,~IEEE,} Xie Chen,~\IEEEmembership{Senior Member,~IEEE,} and Jia Liu,~\IEEEmembership{Member,~IEEE}
\thanks{This work was supported in part by the National Key Research and Development Program of China, under Grant No. 2021YFA1000500(4), in part by the National Natural Science Foundation of China under Grant Nos. 62276153 and U25A20409, in part by SJTU Med-X (Medicine \& Engineering) Translational Research Grant (YG2025LC09). (Corresponding author: Pingyi Fan)}
\thanks{Pingyi Fan, Anbai Jiang, Shuwei Zhang, Wenrui Liang, Junjie Li, Wei-Qiang Zhang and Jia Liu are with the Department of Electronic Engineering, Tsinghua University, Beijing, 100084, P. R. China (e-mail: \{fpy, wqzhang, liuj\}@tsinghua.edu.cn, \{jab22, zhangsw25, lwr24, lijunjie19\}@mails.tsinghua.edu.cn). Wei-Qiang Zhang is also with the Institute for Embodied Intelligence and Robotics, Tsinghua University.}
\thanks{Xinhu Zheng, Bing Han, Yanmin Qian and Xie Chen are with the Department of Computer Science and Engineering, Shanghai Jiao Tong University, Shanghai, 200240, P. R. China (e-mail: \{zhengxh24, hanbing97, yanminqian, chenxie95\}@sjtu.edu.cn).}
\thanks{Zhiqiang Lv was with Huakong AI Plus Company Limited, Beijing, 100084, P. R. China, during the development of this work. He is now with Didi International Business Group, Beijing, P. R. China (e-mail: zedlv@didiglobal.com).}
\thanks{Jia Liu is also with Huakong AI Plus Company Limited, Beijing, 100084, P. R. China (e-mail: liujia@aithu.com).}}

\maketitle

\begin{abstract}

Industrial signal analysis is hindered by severe data heterogeneity, which we characterize as the M5 problem. Existing solutions rely on specialized models that lack robustness and scalability, while large-scale pre-training has rarely been investigated in this area. In this work, we derive a prioritized roadmap for the M5 problem and propose FISHER, a \underline{F}oundation model for multi-modal \underline{I}ndustrial \underline{S}ignal compre\underline{HE}nsive \underline{R}epresentation. To address the foremost multi-sampling-rate problem, FISHER utilizes a novel sub-band modeling approach that treats sampling rate increments as concatenated sub-band information, enabling the adaptive usage of full signal bandwidth without resampling. FISHER is pre-trained by teacher-student self-distillation over external audio and music data. We also establish the RMIS benchmark, comprising 19 datasets across four modalities. In the experiment, FISHER outperforms 24 state-of-the-art series encoders (up to 2B) with much smaller sizes (up to 16x), showcasing groundbreaking diagnostic accuracy and remarkable versatility. We further demonstrate that 1) seamless adaptation to variable sampling rates is the key to generalization 2) audio and music data provide better temporal variability, which is essential for pre-training. Both \href{https://github.com/jianganbai/FISHER}{FISHER} and \href{https://jianganbai.github.io/RMIS}{RMIS} are open-sourced.
\end{abstract}


\section{Introduction}
\label{sec:intro}

\IEEEPARstart{R}{ecent} years saw the emergence of foundation models in multiple domains, such as natural language processing~\cite{xu2025qwen2.5omni}, computer vision~\cite{dosovitskiy2021an,he2022masked,simeoni2025dinov3}, speech~\cite{baevski2020wav2vec,radford2023whisper}, and audio~\cite{huang2022masked,pmlr-v202-chen23ag,chen2024eat}. These models are pre-trained on extensive data and can generalize across various tasks, showcasing groundbreaking capabilities and remarkable versatility even under zero-shot conditions. Motivated by these successes, we aim to answer: \textbf{Is it possible to train a foundation model for all kinds of series-like industrial signals that generalizes across modalities and analysis tasks even without fine-tuning?}

This question is a matter of great concern for both academia and industry. Nowadays, multi-modal industrial signals can be readily acquired thanks to the wide deployment of supervisory control and data acquisition (SCADA) systems. However, how to effectively and efficiently analyze these signals remains challenging. Conventional models are limited to small specific problems, such as sound-based anomaly detection~\cite{jiang2023unsupervised,jiang2024anopatch,wilkinghoff2024self,han2025exploring}, vibration-based bearing fault diagnosis~\cite{wang2023auto,10621043,peng2025bearllm,chen2026cows}, and vibration-based gear fault diagnosis~\cite{8962952,liu2023review,li2025comprehensive}. These models are trained on small-scale datasets, which are more or less deficient in robustness under diverse working conditions. Moreover, one must train a specialized model for each unique combination of modality, machine, and task, which incurs huge burdens during development and deployment. These problems are also concerned by recent studies on foundation models for rotating machinery~\cite{lai2024bearingfm,wang2025rmgpt,li2025hse}. However, these works are limited to vibration signals of bearing and gear, without exploring potential generalization across modalities and machines. More importantly, the ability of generalizing to completely unseen tasks has hardly been discussed previously, which is crucial for foundation models.

Therefore, our target is to advance the frontier of industrial signal foundation models to an unprecedented level of generalization and versatility. To this end, we first summarize key challenges in modeling industrial signals. While pre-training foundation models for speech and audio typically require millions of hours of data, such a volume is no longer considered substantial for industrial signals. We argue that the primary challenge lies in data quality rather than data quantity. Specifically, industrial signals are considerably \textbf{stationary}, yet their patterns are \textbf{heterogeneous} in multiple aspects. In this work, we boil it down to the \textbf{M5} problem:
\begin{enumerate}
    \item \textbf{Multi-modal}. Sound, vibration, voltage, current, etc.
    \item \textbf{Multi-sampling-rate}. The sensor sampling rate is often set as twice the Nyquist bandwidth to reduce cost. Common sampling rates range from 3~kHz to 100~kHz.
    \item \textbf{Multi-scale}. Due to the differences in operating mechanisms (sliding, rotating, static, etc) and working conditions, the signal characteristics are diverse.
    \item \textbf{Multitask}. Anomaly detection, fault diagnosis, remaining useful life (RUL) estimation, etc.
    \item \textbf{Minim fault}. Fault data are often scarce, and the class distribution is often imbalanced.
\end{enumerate}
In that way, industrial signals shall form numerous disjoint small clusters in the data space, each of which corresponds to a unique combination of modality, machine, sensor, working condition, etc. Moreover, due to the stationarity, these clusters may still remain isolated when data volume scales up. Therefore, the data quality of industrial signals themselves is inadequate for pre-training foundation models.

\begin{figure*}[t]
\centering
\includegraphics[width=0.8\linewidth]{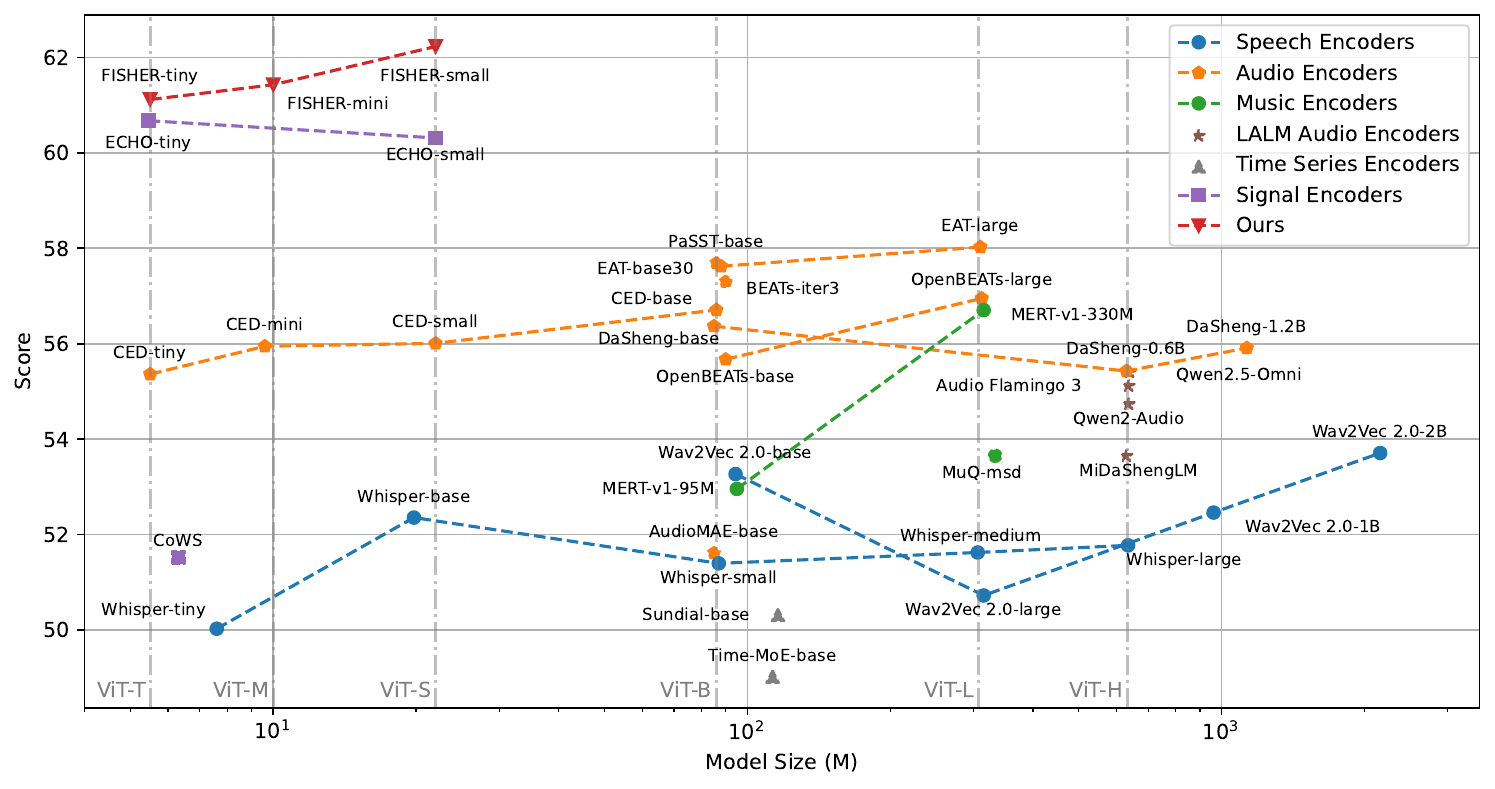}
\caption{Model Performances on the RMIS benchmark (poor models are neglected), where the higher the score is, the better the model is. FISHER achieves the best score with smaller sizes, showcasing superior versatility and scaling properties.}
\label{fig:RMIS_all}
\end{figure*}

However, we argue that the M5 problem is not intractable. We delve into the intrinsic nature of the M5 problem, and combine it with empirical experiences of large-scale pre-training, thereby proposing a prioritized roadmap for solving~it:

\begin{enumerate}
\item \textbf{Special model design for active adaptation to multi-sampling-rate.}
\item \textbf{Large-scale pre-training with high-quality data to smooth multi-modal, multi-scale and multitask.}
\item \textbf{High quality representation for minim fault.}
\end{enumerate}
Detailed analysis is presented in Section~\ref{subsec:pri_roadmap}.

Based on the roadmap, we propose \textbf{FISHER}, short for \textbf{F}oundation model for multi-modal \textbf{I}ndustrial \textbf{S}ignal compre\textbf{HE}nsive \textbf{R}epresentation. FISHER serves as a unified feature extractor for M5 industrial signals and mainly deals with the foremost multi-sampling-rate problem (step 1 of the roadmap). It models the increment of sampling rate as the addition of short-time Fourier transform (STFT) sub-band information. It takes the STFT sub-band as the modeling unit and builds up the signal representation by concatenating sub-band representations. Specifically, a signal is first converted to STFT with fixed time-frequency resolution, then split into sub-bands of fixed bandwidth. FISHER processes these sub-bands independently with a ViT~\cite{dosovitskiy2021an} encoder and sequentially concatenates sub-band representations to form the overall representation. FISHER is pre-trained by a teacher-student self-distillation scheme, where the student is guided by the representations of the teacher, and the teacher is an exponential moving average (EMA) version of the student. Contrary to common belief, FISHER is pre-trained by using audio and music data, which are not directly related to industrial signals. We demonstrate that such data provide similar inductive bias while offering greater temporal variability, making it more beneficial for pre-training than domain-specific industrial signals. To the best of our knowledge, such a kind of foundation model has never been proposed previously.

To comprehensively evaluate the model, we also develop the \textbf{RMIS} benchmark, short for \textbf{R}epresentation of \textbf{M}5 \textbf{I}ndustrial \textbf{S}ignals. The RMIS benchmark incorporates 19 sub-datasets with two typical analysis tasks, i.e. anomaly detection (no fault as prior) and fault diagnosis (identify specific fault types). For fault diagnosis datasets without official split, we adopt sealed train-test split instead of random split to reduce information leakage between train and test sets. All models are evaluated by k-nearest neighbor (KNN) inference without fine-tuning.

On the RMIS benchmark, FISHER is compared with top encoders from speech, audio, music, Large Audio Language Models (LALM) and signal, where FISHER achieves the state-of-the-art (SOTA) performance of 62.23\%, surpassing all baselines by at least 1.56\% and demonstrating remarkable versatility. Meanwhile, FISHER possesses a much more efficient scaling curve, achieving superior performances with much smaller sizes. Based on these results, two key findings can be drawn: 1) the performance gain of FISHER primarily stems from its seamless adaptation to variable sampling rates 2) audio and music data are more effective for pre-training than industrial signals, even though the model is intended for signal analysis. Therefore, our main contributions are:
\begin{itemize}
    \item We formulate key challenges as the M5 problem and provide a prioritized roadmap for solving it.
    \item We propose FISHER\footnote{\url{https://github.com/jianganbai/FISHER}}, the first \textbf{F}oundation model for multi-modal \textbf{I}ndustrial \textbf{S}ignal compre\textbf{HE}nsive \textbf{R}epresentation. FISHER models the STFT sub-bands via teacher-student self-distillation, which enables it to process arbitrary industrial signals without resampling. FISHER achieves SOTA results with much smaller sizes.
    \item We demonstrate that directly handling native high sampling rate is crucial for high-quality signal representation, which has been ignored in previous works.
    \item We point out that an industrial signal foundation model does not necessarily need to be trained on industrial signals and provide novel, critical, and empirical guidelines on data curation for pre-training.
    \item We open-source RMIS\footnote{\url{https://jianganbai.github.io/RMIS}} to benefit the community.
\end{itemize}

\begin{figure}[t]
\centering
\subfloat[16~kHz]{
    \includegraphics[width=0.3\linewidth]{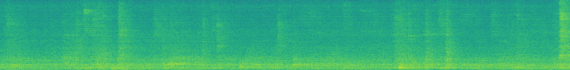}
}
\subfloat[32~kHz]{
    \includegraphics[width=0.3\linewidth]{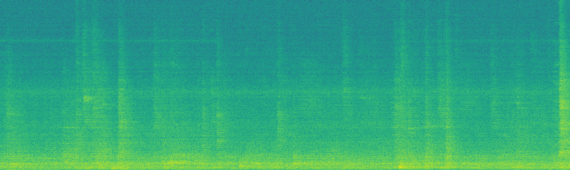}
}
\subfloat[48~kHz]{
    \includegraphics[width=0.3\linewidth]{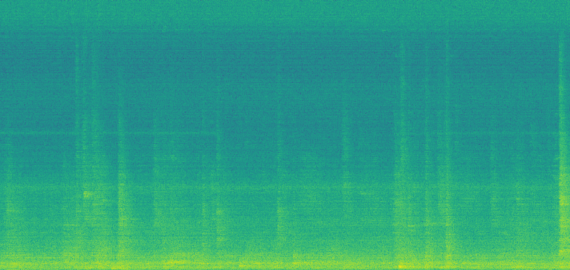}
}
\caption{STFT spectrograms of the same source under different sampling rates. A higher sampling rate comprises additional high-frequency sub-bands with extra information, while its lower sub-bands are almost identical to those of lower sampling rates. Therefore, it is heuristic to take the sub-band as modeling unit.}
\label{fig:spec_msr}
\end{figure}

\section{Related Works}
\label{sec:rela_work}

Industrial signals are continuous series-like data. Thus, one can draw upon relevant experiences from foundational models in speech, audio, and time series. Speech is more variable in temporal, thus speech models~\cite{baevski2020wav2vec,radford2023whisper} usually model the dependencies between time frames. Compared with speech, general audio is much sparser in semantics and lacks frame-level annotations, which more resembles industrial signals. Audio models consider spectrogram patch as the modeling unit and are mostly pre-trained by self-supervised learning (SSL). Typical choices are masked autoencoder~\cite{huang2022masked,dinkel2024scaling}, teacher student self-distillation~\cite{dinkel2023ced,chen2024eat}, and iterative self-tokenization and prediction~\cite{pmlr-v202-chen23ag,bharadwaj2025openbeats}. With the rise of LALM~\cite{chu2024qwen2,xu2025qwen2.5omni,ghosh2026audio,dinkel2025midashenglm}, these speech and audio models are utilized as the audio encoders to large language models (LLM) and are fine-tuned on high-quality audio-text pairs. Compared with speech and audio models, time series models~\cite{shi2025time,liu2025sundial} focus on series with much smaller scale (thousands of points) and much less noise. These models generally take in raw data points without Fourier transform, and are trained to forecast future time steps. Thus, audio models are better aligned with our requirements.


Meanwhile, large-scale pre-training is gradually revolutionizing industrial signal analysis tasks. For anomalous sound detection (ASD), fine-tuning audio pre-trained models has become the dominant approach~\cite{jiang2024anopatch,han2025exploring}. For vibration-based fault diagnosis, transferring image pre-trained backbones was once the research hotspot~\cite{li2022cstt}. Some recent works adopt off-the-shelf SSL schemes for pre-training, such as contrastive learning~\cite{lai2024bearingfm} and autoregression~\cite{wang2025rmgpt}. Other works transfer LLM~\cite{pang2024parinfogpt} or time series foundation models~\cite{li2026data} to forecast the signal series. There are also works on building LALM-like models~\cite{peng2025bearllm} for text-form fault analysis. While these works have successfully applied the pre-training paradigm to signal analysis, none of them focus on extracting general-purpose features for multi-modal industrial signals.


\section{FISHER}
\label{sec:model}

\subsection{Prioritized Roadmap}
\label{subsec:pri_roadmap}

We believe large-scale pre-training is the key to the M5 problem. We now analyze how it separately breaks down the M5 problem and gradually derive the solution roadmap.

\subsubsection{High quality representation for minim fault}

Recent breakthrough in ASD~\cite{jiang2024anopatch,jiang2025adaptive} suggests that as long as the representation quality is good, one can detect anomaly even without prior knowledge. Thus, the minim fault problem does not present any difficulty for representation learning.

\subsubsection{Large-scale pre-training with high-quality data to smooth multi-modal, multi-scale and multitask}

We believe the M5 problem is mostly related to the appearance (how it is perceived) rather than the essence (what it conveys). In fact, there are some internal similarities beneath the M5 problem:

%
\begin{itemize}
    \item Sound and vibration are essentially two forms of vibration, since sound travels through the air as vibration.
    \item Different signals are perceptions of the same mechanical event by different physical laws.
    \item Fault patterns are comparable since machines are assembled from shared components.
    \item Most downstream tasks are related to malfunctions. A dense representation is sufficient for multiple tasks.
\end{itemize}
These similarities can be covered and utilized through large-scale pre-training, and a single foundation model can generalize to heterogeneous data and tasks, as has been demonstrated across various domains. However, to derive that model, training solely on industrial signals is insufficient due to the poor data quality. We contend that the pre-training data should primarily consist of high-quality data with similar inductive bias to industrial signals. Section~\ref{subsec:data_select} further demonstrates that audio and music data are good selections for this work.

\subsubsection{Special model design for active adaptation to multi-sampling-rate}

The multi-sampling-rate problem is purely a matter of the appearance. Since sensors are designed to satisfy the Nyquist Sampling Theorem, the information is fully retained in the original signal. However, most prior works trained models with a fixed sampling rate, which leads to huge information loss during resampling. This not only leads to insufficient learning during pre-training, but also limits its understanding of critical high-frequency information during inference. Section~\ref{subsec:high_freq_gain} demonstrates the importance of high-sampling-rate adaptation. To effectively conduct large-scale pre-training, the model must be designed to accept input~with arbitrary sampling rates. Thus, the multi-sampling-rate problem is the foremost problem that must be resolved.

\subsection{Sub-band Modeling}
\label{subsec:sb_model}

FISHER innovatively adopts an STFT sub-band modeling scheme to address the multi-sampling-rate problem. To capture long-range dependencies, FISHER accepts inputs up to 10~seconds, which is significantly larger than specialized fault diagnosis models. To reduce redundancy, FISHER converts the signal to STFT, which is contrary to the commonly-adopted log-mel spectrogram in speech and audio models, since:
\begin{itemize}
    \item Malfunctions often appear in high frequencies, which would be diluted in mel scale.
    \item The harmonic relationships of frequencies are vital, which would be smoothed in mel scale.
\end{itemize}
To deal with the multi-sampling-rate problem, the STFT window size $N$ is mapped to fixed time duration $t_{win}$: Let $f_s$ denote the signal sampling rate, then $N=t_{win}{\cdot}f_s$. Thus, the frequency resolution will be constant for arbitrary $f_s$:
\begin{equation}
    {\Delta}f=\frac{f_s}{N}=\frac{f_s}{t_{win}{\cdot}f_s}=\frac{1}{t_{win}}
\end{equation}
where ${\Delta}f$ is the frequency gap between adjacent frequency grids. Similarly, the hop size is also mapped to fixed time duration $t_{hop}$, such that the time resolution is constant, and signals with the same time duration will have spectrograms with the same time shape regardless of the sampling rate.

\begin{figure*}[t]
\centering
\includegraphics[width=0.75\textwidth]{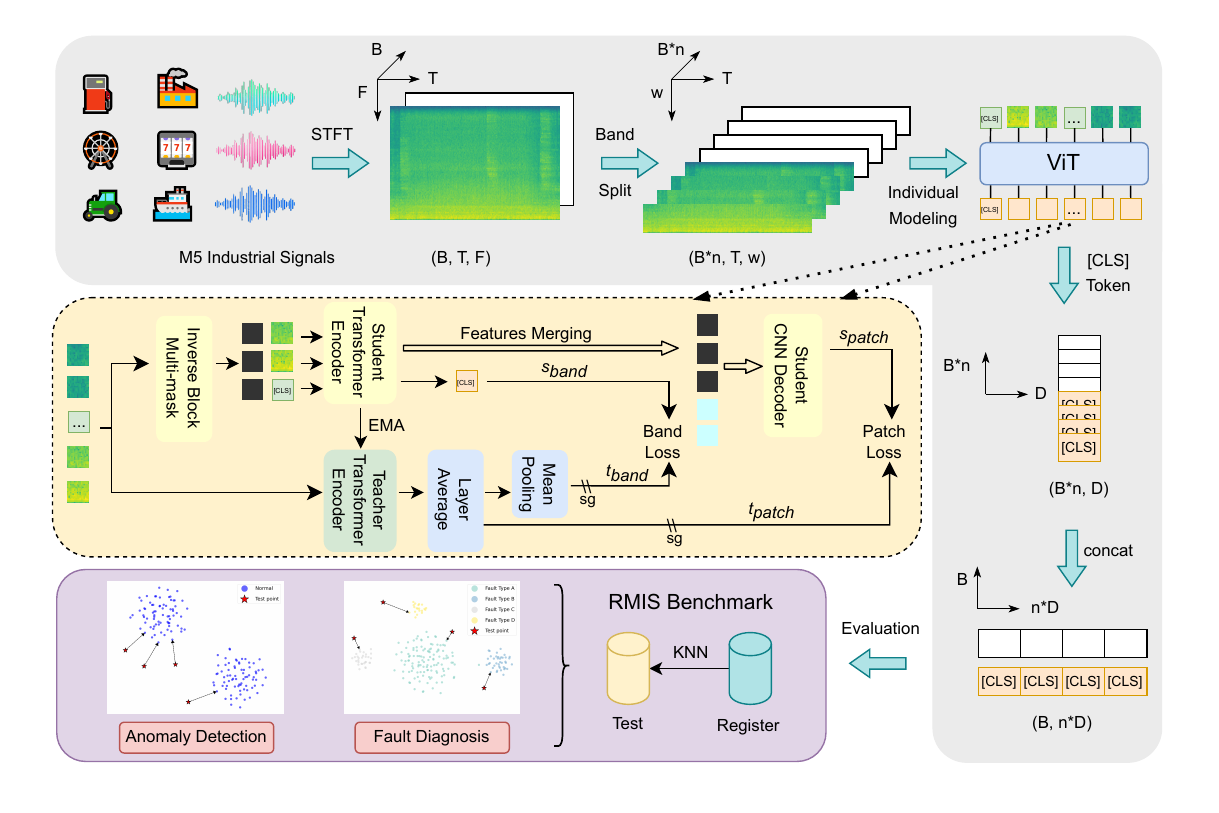}
\caption{Pipeline of FISHER and RMIS. FISHER converts signals into STFT spectrograms and splits them into sub-bands with fixed bandwidth $w$. These sub-bands are processed individually by the ViT backbone, and the [CLS] embeddings are concatenated as the signal representations. The ViT backbone is trained by teacher-student self-distillation, where the teacher encoder is an EMA version of the student encoder. The model is evaluated on the RMIS benchmark by KNN inference.}
\label{fig:bs_pipe}
\end{figure*}

To deal with the variable frequency shape, FISHER emphasizes the importance of sub-band and considers it as the building blocks of the overall information. On the one hand, as depicted in Fig.~\ref{fig:spec_msr}, the information gain of a higher sampling rate lies in the additional sub-bands. As is known, all sensors employ anti-aliasing filtering to prevent signal aliasing. Therefore, the spectrogram does not contain any information about frequencies higher than half the sampling rate. On the other hand, sampling rates of common large-scale datasets, i.e. 16~kHz, 32~kHz, 44.1~kHz, and 48~kHz, are integer multiples of a fundamental frequency $f_{base}$, such as 2~kHz and 4~kHz, making sub-band a natural unit for modeling multi-sampling-rate signals. We thus take the sub-band as the modeling unit and build up the information of the whole spectrogram by concatenating sub-band information just like building blocks. That is, the signal representation is the concatenation of sub-band representations. The higher the sampling rate is, the more informative the representation is.

We now describe the sub-band modeling process in detail, which is illustrated in Fig.~\ref{fig:bs_pipe}. A batch of signals is first resampled to a batch-specific sampling rate $sr_{batch}$ to align the spectrogram shape for batching, where $sr_{batch}$ is randomly selected from all the harmonics of $f_{base}$ that are smaller than a maximum frequency $f_{max}$. This augmentation method prompts the model to better learn the similarities and differences between sampling rates during pre-training, thereby allowing it to directly process raw sampling rates during inference. The aligned signals are then converted to log amplitude STFT spectrograms of shape $(B,\ T,\ F)$, where $B$ is the batch size, and $T$ and $F$ are the time and frequency shapes respectively. Each spectrogram is then split into sub-bands with bandwidth $w$ and concatenated along the batch axis, thereby transferring the variability from the frequency axis to the batch axis. These sub-bands have a shape of $(B{\times}n,\ T,\ w)$, where $w=f_{base}{\cdot}t_{win}$ is the sub-band bandwidth, and $n=\lfloor\frac{F}{w}\rfloor$ is the number of sub-bands. They are then processed individually by the network, and their representations are concatenated afterwards.

\begin{figure*}[t]
\centering
\includegraphics[width=0.9\textwidth]{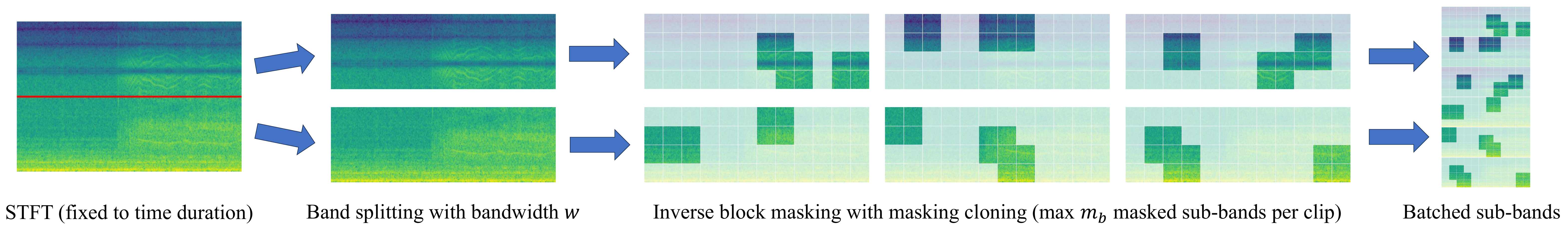}
\caption{Pipeline of band splitting, inverse masking and mask cloning.}
\label{fig:spec_mask}
\end{figure*}

\subsection{SSL Pre-training}
\label{subsec:model_arch}

FISHER adopts a teacher-student self-distillation SSL framework. It comprises three sub-networks: a student encoder~$E_{stu}$, a student decoder~$D_{stu}$ and a teacher encoder~$E_{tea}$. $E_{tea}$ is an exponential moving average (EMA) version of $E_{stu}$. Specifically, they have the identical ViT~\cite{dosovitskiy2021an} structure with fixed sinusoidal position encoding and post-norm, while their parameters are correlated by EMA:
\begin{equation}
    \theta_{E_{tea}}^{(t+1)}=\tau\theta_{E_{tea}}^{(t)}+(1-\tau)\theta_{E_{stu}}^{(t)}
    \label{eq:ema}
\end{equation}
where $\theta_{E_{tea}}^{(t)}$ and $\theta_{E_{stu}}^{(t)}$ are the parameters of $E_{tea}$ and $E_{stu}$ at step $t$ respectively, and $\tau$ is the EMA decay factor.

The training pipeline of FISHER is illustrated in Fig.~\ref{fig:bs_pipe}. We employ a masking modeling scheme, where $E_{stu}$ processes only the unmasked parts of sub-bands, while $E_{tea}$ has access to the complete sub-band. Thus, the representation quality of $E_{stu}$ is inferior to that of $E_{tea}$, which provides a clear optimization direction: $E_{stu}$ is trained to align with the representation of $E_{tea}$, while $E_{tea}$ is detached from gradient map and is updated by Eq.~\ref{eq:ema} per step. This enables the model to learn through self-distillation on large amount of unlabeled signals. Specifically, the representation alignment is conducted on both sub-band and patch levels, and $D_{stu}$ is introduced to facilitate the patch-level alignment.

We now introduce the SSL scheme in detail. For the student branch, inverse block masking is first applied on the sub-bands, which is illustrated in Fig.~\ref{fig:spec_mask}. We randomly select patch blocks with size $m_l$ and discard unselected patches. Mask cloning is also employed, where we apply $\lfloor\frac{m_b}{n}\rfloor$ masks for each sub-band to generate multiple views of the same sub-band. Here, $m_b$ is the maximum number of views sourced from the same clip, and we constrain it to prevent GPU load imbalance. Since these views can reuse the same teacher representation, mask cloning equivalently increases the batch size with less computation and I/O overhead. Unmasked patches of each sub-band are then flattened into a patch sequence, appended with a [CLS] token at the front and encoded by $E_{stu}$. The outputs of $E_{stu}$ are then converted to representations of two levels. Specifically, the [CLS] token output is selected as the student sub-band representation $s_{band}$ with size $(B{\times}n,\ d)$, where $d$ is the hidden size of the Transformer block. Meanwhile, the outputs of unmasked patches are merged with the masked parts in the original spatial locations, where the values of the masked parts are initialized from normal Gaussian. The merged sequence is further passed to $D_{stu}$, and its output is selected as the student patch representation $s_{patch}$ with size $(B{\times}n,\ L,\ d)$, where $L$ is the total number of patches.

For the teacher branch, $E_{tea}$ processes the complete sub-band in the same ViT procedure, which splits each sub-band into patches and processes the patch sequence, while masking is not applied. The embeddings of all layers are first averaged across layers to form the teacher patch representation $t_{patch}$ with size $(B{\times}n,\ L,\ d)$, and $t_{patch}$ is further averaged across patches to form the teacher sub-band representation $t_{band}$ with size $(B{\times}n,\ d)$. Finally, self-distillation is conducted on both sub-band and patch levels:
\begin{equation}
    \begin{cases}
        &L_{band}=\|s_{band}-sg(t_{band})\|_2^2\\
        &L_{patch}=\|s_{patch}^{(mask)}-sg(t_{patch}^{(mask)})\|_2^2
    \end{cases}
\end{equation}
where the $(mask)$ superscript selects all masked patches, and $sg(\cdot)$ denotes stop gradient. The final loss is a linear combination of the two losses with coefficient $\lambda$:
\begin{equation}
    L=\lambda{\cdot}L_{band}+(1-\lambda){\cdot}L_{patch}
\end{equation}

During inference, only $E_{stu}$ is employed. Spectrograms are split into sub-bands and processed by $E_{stu}$ without masking. The final representation is the concatenation of all $s_{band}$.

\begin{table}[t]
\centering
\caption{Key features of datasets in the RMIS benchmark. Split denotes whether an official train-test split is provided.}
\label{tab:rmis_key_feat}
\resizebox{\linewidth}{!}{
\begin{tabular}{*{7}{c}}
\toprule
\multirow{2}*{Task} & \multirow{2}*{Dataset} & \multirow{2}*{Modality} & Num & Sampling & \multirow{2}*{Volume} & \multirow{2}*{Split} \\
& & & Class & Rate & & \\
\midrule
\multirow{6}*{\makecell[c]{Anomaly\\Detection}} & DCASE20 & Sound & 2 & 16~kHz & 153 h & \checkmark \\
& DCASE21 & Sound & 2 & 16~kHz & 165 h & \checkmark \\
& DCASE22 & Sound & 2 & 16~kHz & 139 h & \checkmark \\
& DCASE23 & Sound & 2 & 16~kHz & 50 h & \checkmark \\
& DCASE24 & Sound & 2 & 16~kHz & 49 h & \checkmark \\
& DCASE25 & Sound & 2 & 16~kHz & 45 h & \checkmark \\
\midrule
\multirow{13}*{\makecell[c]{Fault\\Diagnosis}} & IICA & Sound & 6 & 48~kHz & 47 h & \ding{55} \\
& IIEE & Sound & 3 & 44.1~kHz & 1 h & \checkmark \\
& WTPG & Vibration & 5 & 48~kHz & 14 h & \ding{55} \\
& MaFaulDa\_sound & Sound & 10 & 50~kHz & 3 h & \ding{55} \\
& MaFaulDa\_vib & Vibration & 10 & 50~kHz & 16 h & \ding{55} \\
& SDUST\_bearing & Vibration & 10 & 25.6~kHz & 25 h & \ding{55} \\
& SDUST\_gear & Vibration & 7 & 25.6~kHz & 17 h & \ding{55} \\
& UMGED\_sound & Sound & 11 & 51.2~kHz & 59 h & \ding{55} \\
& UMGED\_vib & Vibration & 11 & 51.2~kHz & 176 h & \ding{55} \\
& UMGED\_vol & Voltage & 11 & 51.2~kHz & 117 h & \ding{55} \\
& UMGED\_cur & Current & 11 & 51.2~kHz & 117 h & \ding{55} \\
& PU\_vib & Vibration & 3 & 64~kHz & 3 h & \ding{55} \\
& PU\_cur & Current & 3 & 64~kHz & 6 h & \ding{55} \\
\bottomrule
\end{tabular}}
\end{table}

\section{RMIS Benchmark}
\label{sec:bench}

To evaluate the comprehensive representation ability of the model, we develop the RMIS benchmark, which comprises six anomaly detection datasets and 13 fault diagnosis datasets, whose key features are presented in Table~\ref{tab:rmis_key_feat}. Let $S_{AD}$ denote the mean detection score of anomaly detection datasets, which is based on the area under the receiver operating characteristic curve (AUC). Let $S_{FD}$ denote the mean accuracy of fault diagnosis datasets. The overall RMIS score $S_{RMIS}$ is the mean of $S_{AD}$ and $S_{FD}$. To emphasize the emergent versatility of foundation models, all models are evaluated by KNN inference without any fine-tuning. We briefly introduce these datasets below, and details are presented in Appendix~\ref{appsec:rmis_detail}. 

\newcounter{admc}
\newcounter{fdmc}
\newcounter{allc}
\setcounter{admc}{0}
\setcounter{fdmc}{0}
\setcounter{allc}{0}

\begingroup
\setlength{\tabcolsep}{3.5pt}  
\renewcommand{\arraystretch}{1.4}  
\begin{table*}[t]
\centering
\caption{Results on the RMIS Benchmark ($\uparrow$)}
\label{tab:RMIS_num}
\resizebox{\textwidth}{!}{%
\begin{threeparttable}
\begin{tabular}{
*{9}{c}
>{\stepcounter{admc}\ifnum\value{admc}>2 \cellcolor{gray!15}\fi}c
*{13}{c}
>{\stepcounter{fdmc}\ifnum\value{fdmc}>2 \cellcolor{gray!15}\fi}c
>{\stepcounter{allc}\ifnum\value{allc}>3 \cellcolor{gray!15}\fi}c
}
\toprule
\multirow{3}*[-1.6ex]{Model} & \multirow{3}*[-1.6ex]{Variant} & \multirow{3}*[-1.6ex]{Size} & \multicolumn{7}{c}{Anomaly Detection} & \multicolumn{14}{c}{Fault Diagnosis} & \multirow{3}*[-1.6ex]{$\bm{S_{RMIS}}$} \\
\cmidrule(lr){4-10} \cmidrule(lr){11-24}
 & & & \multicolumn{6}{c}{DCASE} & \multirow{2}*[-0.4ex]{$\bm{S_{AD}}$} & \multirow{2}*[-0.4ex]{IICA} & \multirow{2}*[-0.4ex]{IIEE} & \multirow{2}*[-0.4ex]{WTPG} & \multicolumn{2}{c}{MaFaulDa} & \multicolumn{2}{c}{SDUST} & \multicolumn{4}{c}{UMGED} & \multicolumn{2}{c}{PU} & \multirow{2}*[-0.4ex]{$\bm{S_{FD}}$} & \\
\cmidrule(lr){4-9} \cmidrule(lr){14-15} \cmidrule(lr){16-17} \cmidrule(lr){18-21} \cmidrule(lr){22-23}
 & & & 20 & 21 & 22 & 23 & 24 & 25 & & & & & sound & vib & bearing & gear & sound & vib & vol & cur & vib & cur & & \\
\midrule
\multirow{4}*{\textcolor{mpblue}{\textbf{Wav2Vec 2.0}}} & base & 95M & 64.76 & 55.14 & 55.77 & 55.69 & 52.76 & 55.69 & 56.64 & 46.77 & 96.61 & 60.81 & 60.72 & 78.34 & 54.12 & 95.01 & 9.08 & 9.34 & 11.93 & 15.58 & 67.00 & 43.42 & 49.90 & 53.27 \\
 & large & 315M & 65.61 & 55.23 & 54.44 & 48.09 & 50.51 & 33.98 & 51.31 & 45.49 & 72.48 & 78.33 & 62.83 & 87.35 & 49.98 & 97.04 & 10.12 & 8.40 & 9.18 & 12.50 & 69.24 & 48.86 & 50.14 & 50.72 \\
 & 1B & 962M & 65.50 & 55.22 & 55.81 & 54.12 & 51.83 & 51.60 & 55.68 & 44.23 & 94.97 & 84.84 & 49.40 & 77.08 & 49.95 & 96.72 & 8.39 & 10.89 & 8.49 & 9.93 & 64.59 & 40.69 & 49.24 & 52.46 \\
 & 2B & 2.16B & 65.49 & 55.47 & 56.35 & 55.47 & 53.91 & 54.53 & 56.87 & 42.09 & 78.81 & 84.49 & 59.96 & 84.14 & 50.31 & 98.24 & 8.37 & 12.06 & 10.80 & 11.13 & 72.67 & 44.07 & 50.55 & 53.71 \\
\spacedashline
\multirow{5}*{\textcolor{mpblue}{\textbf{Whisper}}} & tiny & 7.6M & 62.85 & 54.42 & 53.72 & 54.09 & 52.53 & 54.01 & 55.27 & 40.36 & 55.16 & 71.86 & 44.40 & 76.92 & 51.17 & 96.99 & 6.71 & 9.52 & 9.48 & 12.51 & 66.00 & 41.12 & 44.78 & 50.03 \\
 & base & 19.8M & 65.66 & 54.89 & 56.12 & 56.90 & 53.70 & 56.14 & 57.24 & 40.10 & 77.24 & 83.95 & 36.35 & 80.82 & 51.30 & 96.90 & 8.98 & 9.71 & 9.02 & 10.21 & 69.95 & 42.70 & 47.48 & 52.36 \\
 & small & 87M & 64.02 & 54.44 & 54.85 & 55.50 & 53.79 & 54.67 & 56.21 & 41.12 & 70.21 & 83.55 & 38.72 & 76.75 & 49.23 & 95.84 & 8.63 & 10.50 & 8.68 & 9.89 & 68.89 & 43.57 & 46.58 & 51.40 \\
 & medium & 306M & 65.41 & 54.76 & 55.58 & 56.02 & 54.13 & 55.21 & 56.85 & 41.28 & 66.84 & 85.62 & 37.31 & 76.66 & 48.93 & 96.11 & 9.29 & 11.79 & 8.09 & 9.61 & 67.65 & 44.02 & 46.40 & 51.63 \\
 & large-v3 & 635M & 62.27 & 53.65 & 54.51 & 53.97 & 52.53 & 54.54 & 55.25 & 47.57 & 70.44 & 87.24 & 59.32 & 73.36 & 50.85 & 94.08 & 9.14 & 10.10 & 8.34 & 8.05 & 68.80 & 40.68 & 48.31 & 51.78 \\
\midrule
\textcolor{mporange}{\textbf{PaSST}} & base & 86M & 66.60 & 56.84 & 57.58 & 59.56 & 57.64 & 57.76 & 59.33 & 60.50 & 97.56 & \bstscr{92.69} & 59.30 & 88.59 & 61.86 & 97.11 & 7.16 & 9.80 & 10.64 & 15.87 & 74.92 & 52.62 & 56.05 & 57.69 \\
\spacedashline
\textcolor{mporange}{\textbf{AudioMAE}} & base & 85M & 63.87 & 54.86 & 56.34 & 57.96 & 55.30 & 56.82 & 57.52 & 58.88 & 53.89 & 64.02 & 47.48 & 80.95 & 59.74 & 98.18 & 6.91 & 10.93 & 10.01 & 11.59 & 56.02 & 35.54 & 45.70 & 51.61 \\
\spacedashline
\textcolor{mporange}{\textbf{BEATs}} & iter3 & 90M & \bstscr{74.20} & \bstscr{60.50} & 58.95 & 63.31 & 56.93 & 59.09 & \bstscr{62.16} & 77.93 & 68.17 & 70.57 & 52.87 & 84.45 & 62.37 & 98.23 & 5.98 & 10.32 & 10.95 & 17.23 & 74.18 & 48.42 & 52.44 & 57.30 \\
\spacedashline
\multirow{2}{*}{\textcolor{mporange}{\textbf{OpenBEATs}}} & base & 90M & 70.86 & 58.60 & 57.45 & 62.48 & 56.10 & 57.32 & 60.47 & 77.87 & 70.93 & 63.60 & 45.50 & 83.39 & 57.31 & 96.51 & 5.47 & 9.51 & 11.09 & 13.78 & 76.96 & 49.50 & 50.88 & 55.67 \\
 & large & 310M & 72.78 & 59.93 & 58.29 & 63.29 & 57.32 & 58.03 & 61.61 & 75.18 & 82.86 & 62.33 & 42.69 & 79.37 & 70.65 & 96.86 & 6.91 & 9.66 & 12.63 & 12.48 & \bstscr{78.57} & 49.67 & 52.30 & 56.95 \\
\spacedashline
\multirow{2}{*}{\textcolor{mporange}{\textbf{EAT}}} & base30 & 88M & 72.69 & 58.03 & 58.84 & 59.35 & 56.48 & 58.65 & 60.67 & 64.48 & 96.65 & 61.28 & 68.32 & 91.90 & 64.03 & 95.46 & 7.89 & 9.39 & 12.49 & 18.87 & 68.33 & 50.38 & 54.57 & 57.62 \\
 & large10 & 310M & 73.81 & 57.06 & 57.94 & 60.29 & 57.78 & \bstscr{60.34} & 61.20 & 65.80 & 94.06 & 68.91 & 58.47 & 88.42 & 62.64 & 96.32 & 9.57 & 11.60 & 11.22 & 15.54 & 75.74 & 54.73 & 54.85 & 58.03 \\
\spacedashline
\multirow{4}{*}{\textcolor{mporange}{\textbf{CED}}} & tiny & 5.5M & 66.99 & 56.17 & 56.75 & 60.09 & 56.40 & 58.40 & 59.15 & 47.93 & 74.23 & 89.43 & 53.77 & 84.56 & 59.02 & 96.06 & 8.09 & 12.96 & 10.31 & 12.65 & 71.50 & 49.95 & 51.57 & 55.36 \\
 & mini & 9.6M & 67.48 & 56.35 & 56.59 & 60.05 & 57.44 & 58.26 & 59.36 & 50.21 & 84.26 & 89.26 & 56.95 & 84.88 & 59.24 & 95.50 & 8.36 & 11.66 & 10.13 & 12.43 & 71.50 & 48.56 & 52.53 & 55.95 \\
 & small & 22M & 67.50 & 56.65 & 56.87 & 60.76 & 57.88 & 58.15 & 59.63 & 50.49 & 80.23 & 91.20 & 54.67 & 84.67 & 58.00 & 96.12 & 8.05 & 12.15 & 10.26 & 13.14& 72.89 & 49.08 & 52.38 & 56.01 \\
 & base & 86M & 67.60 & 56.67 & 56.95 & 60.99 & \bstscr{57.89} & 58.45 & 59.76 & 52.26 & 90.08 & 89.20 & 57.64 & 85.95 & 60.20 & 95.84 & 8.12 & 11.57 & 10.03 & 13.63 & 72.14 & 50.91 & 53.66 & 56.71 \\
\spacedashline
\multirow{3}{*}{\textcolor{mporange}{\textbf{DaSheng}}} & base & 85M & 69.25 & 57.28 & 56.02 & 60.95 & 56.39 & 55.82 & 59.28 & 74.42 & 99.10 & 46.84 & 63.53 & 92.06 & 61.37 & 98.30 & 5.64 & 4.45 & 11.74 & 20.08 & 74.93 & 42.42 & 53.45 & 56.37 \\
 & 0.6B & 630M & 68.35 & 56.76 & 55.35 & 59.85 & 56.34 & 55.41 & 58.68 & 75.14 & 99.25 & 45.41 & 57.26 & 91.11 & 58.23 & 97.59 & 5.27 & 4.28 & 11.69 & 14.68 & 75.78 & 42.63 & 52.18 & 55.43 \\
 & 1.2B & 1130M & 69.52 & 57.06 & 55.82 & 61.14 & 56.32 & 56.01 & 59.31 & 73.18 & 99.00 & 48.89 & 58.16 & 90.79 & 55.50 & 97.99 & 5.42 & 4.52 & 11.76 & 17.32 & 75.99 & 44.11 & 52.51 & 55.91 \\
\midrule
\multirow{2}{*}{\textcolor{mpgreen}{\textbf{MERT}}} & v1-95M & 95M & 65.54 & 55.40 & 54.68 & 54.46 & 53.37 & 52.74 & 56.03 & 58.59 & 98.07 & 54.93 & 42.47 & 77.55 & 57.92 & 98.87 & 9.26 & 4.42 & 11.55 & 16.38 & 75.47 & 43.01 & 49.88 & 52.96 \\
 & v1-330M & 315M & 68.79 & 56.87 & 55.50 & 58.34 & 55.17 & 55.65 & 58.39 & 64.11 & 99.85 & 69.02 & 66.88 & 92.38 & 61.71 & 99.19 & 5.54 & 3.92 & 15.02 & 18.57 & 72.97 & 46.12 & 55.02 & 56.70 \\
\spacedashline
\textcolor{mpgreen}{\textbf{MuQ}} & msd & 333M & 66.91 & 56.75 & 56.04 & 57.69 & 55.64 & 56.09 & 58.19 & 62.23 & 95.33 & 37.98 & 41.22 & 88.08 & 60.67 & 96.38 & 6.11 & 5.72 & 15.07 & 9.00 & \bstscr{78.57} & 42.20 & 49.12 & 53.65 \\
\midrule
\textcolor{mpbrown}{\textbf{Qwen2-Audio}} & - & 637M & 67.32 & 55.38 & 56.46 & 57.19 & 56.07 & 56.27 & 58.12 & 45.59 & 73.56 & 89.91 & 59.63 & 83.35 & 60.13 & 97.12 & 7.63 & 10.26 & 9.17 & 10.63 & 73.17 & 47.46 & 51.35 & 54.73 \\
\spacedashline
\textcolor{mpbrown}{\textbf{Qwen2.5-Omni}} & - & 640M & 70.94 & 57.39 & 57.20 & 59.14 & 56.83 & 57.05 & 59.76 & 45.49 & 66.56 & 90.73 & 53.52 & 86.21 & 60.49 & 97.60 & 9.02 & 11.35 & 9.25 & 11.87 & 74.84 & 46.74 & 51.05 & 55.40 \\
\spacedashline
\textcolor{mpbrown}{\textbf{Audio Flamingo 3}} & - & 637M & 67.02 & 55.06 & 57.08 & 57.52 & 56.76 & 56.55 & 58.33 & 47.41 & 88.57 & 92.41 & 40.85 & 87.20 & 58.27 & 98.09 & 7.69 & 11.56 & 9.30 & 10.21 & 73.37 & 49.77 & 51.90 & 55.12 \\
\spacedashline
\textcolor{mpbrown}{\textbf{MiDaShengLM}} & - & 630M & 67.10 & 55.46 & 56.10 & 58.70 & 56.15 & 56.03 & 58.26 & 63.78 & 59.63 & 86.39 & 35.69 & 76.94 & 68.31 & 98.25 & 7.75 & 15.54 & 9.20 & 7.29 & 67.47 & 41.28 & 49.04 & 53.65 \\
\midrule
\textcolor{mpgray}{\textbf{Time-MoE}} & base & 113M & 55.43 & 51.83 & 53.72 & 51.38 & 50.35 & 50.58 & 52.22 & 56.84 & 77.21 & 76.31 & 47.83 & 73.75 & 53.92 & 92.52 & 5.29 & 3.81 & 12.71 & 6.52 & 56.67 & 32.52 & 45.84 & 49.03 \\
\spacedashline
\textcolor{mpgray}{\textbf{Sundial}} & base & 116M & 63.87 & 54.50 & 53.15 & 49.06 & 51.84 & 52.17 & 54.10 & 50.53 & 98.41 & 55.57 & 15.56 & 59.94 & 45.01 & 86.57 & 5.22 & 4.54 & \bstscr{37.30} & 26.10 & 68.73 & 51.52 & 46.54 & 50.32 \\
\midrule
\textcolor{mppurple}{\textbf{PEEMD}} & - & - & 58.84 & 53.25 & 53.32 & 53.88 & 53.63 & 50.63 & 53.93 & 35.40 & 66.41 & 61.29 & 51.23 & 54.30 & 30.45 & 53.14 & 10.85 & 10.55 & 14.58 & 10.69 & 63.64 & 36.49 & 38.39 & 46.16 \\
\spacedashline
\textcolor{mppurple}{\textbf{TFPred}} & wc1 & 3.9M & 58.41 & 53.22 & 51.39 & 51.08 & 50.66 & 51.14 & 52.65 & 35.35 & 77.15 & 24.91 & 10.94 & 36.30 & 26.29 & 56.56 & 9.71 & 9.53 & 22.48 & 10.80 & 51.52 & 33.02 & 31.12 & 41.88 \\
\spacedashline
\textcolor{mppurple}{\textbf{LiConvFormer}} & - & 320K & 61.65 & 53.35 & 51.96 & 49.50 & 51.72 & 50.62 & 53.13 & 33.13 & 75.10 & 34.08 & 13.56 & 47.15 & 34.00 & 71.24 & 7.52 & 6.78 & 25.68 & 11.92 & 56.60 & 36.23 & 34.85 & 43.99 \\
\spacedashline
\textcolor{mppurple}{\textbf{BearLLM}} & - & 204K & 64.12 & 53.12 & 48.95 & 44.04 & 51.96 & 50.47 & 52.11 & 26.29 & 66.25 & 65.76 & 45.54 & 48.99 & 44.11 & 77.84 & 12.66 & 9.66 & 11.33 & 5.59 & 65.34 & 32.83 & 39.40 & 45.75 \\
\spacedashline
\textcolor{mppurple}{\textbf{RotLLM}} & - & 770K & 62.91 & 53.55 & 53.80 & 53.63 & 51.16 & 50.94 & 54.33 & 56.98 & 85.82 & 48.33 & 15.65 & 38.65 & 60.10 & 92.24 & 6.89 & 8.13 & 9.36 & 5.00 & 69.79 & 37.02 & 41.07 & 47.70 \\
\spacedashline
\textcolor{mppurple}{\textbf{CoWS}} & - & 6.31M & 59.98 & 53.93 & 53.12 & 52.39 & 52.07 & 50.03 & 53.59 & 44.72 & 74.40 & 86.45 & \bstscr{80.43} & 76.17 & 48.40 & 92.96 & 10.68 & 4.79 & 21.04 & 9.29 & 59.79 & 33.80 & 49.46 & 51.52 \\
\spacedashline
\multirow{2}{*}{\textcolor{mppurple}{\textbf{ECHO}}} & tiny & 5.5M & 70.43 & 59.01 & 58.86 & 63.92 & 55.77 & 56.55 & 60.76 & 90.31 & \bstscr{100.00} & 65.81 & 58.84 & 92.26 & \bstscr{76.70} & 98.78 & 12.90 & 21.34 & 19.92 & 25.77 & 74.21 & 50.87 & 60.59 & 60.67 \\
 & small & 22M & 72.50 & 60.20 & 59.13 & \bstscr{64.25} & 56.76 & 57.45 & 61.71 & \bstscr{90.34} & 99.95 & 60.45 & 61.81 & 93.18 & 69.08 & 98.54 & \bstscr{14.17} & 16.96 & 14.90 & 18.34 & 75.70 & 52.37 & 58.91 & 60.31 \\
\midrule
\multirow{3}{*}{\textcolor{mpred}{\textbf{FISHER}}} & tiny & 5.5M & 70.86 & 58.76 & 56.40 & 58.62 & 53.64 & 56.37 & 59.11 & 85.31 & 98.72 & 83.84 & 75.79 & 92.58 & 71.81 & \bstscr{99.20} & 11.90 & \bstscr{23.58} & 21.10 & 28.24 & 72.36 & \bstscr{56.21} & 63.13 & 61.12 \\
 & mini & 10M & 70.19 & 58.40 & 57.62 & 61.07 & 54.59 & 55.75 & 59.60 & 86.02 & 99.05 & 84.74 & 75.74 & 93.20 & 70.09 & 98.55 & 13.69 & 20.52 & 22.95 & \bstscr{29.97} & 72.01 & 55.73 & 63.25 & 61.43 \\
 & small & 22M & 71.04 & 59.48 & \bstscr{59.64} & 62.63 & 55.62 & 58.46 & 61.15 & 90.23 & 98.96 & 86.77 & 72.61 & \bstscr{95.86} & 76.35 & 99.08 & 12.72 & 18.35 & 17.42 & 25.52 & 74.90 & 54.29 & \bstscr{63.31} & \bstscr{62.23} \\
\bottomrule
\end{tabular}%
\begin{tablenotes}
\item Different colors denote \textcolor{mpblue}{speech encoders}, \textcolor{mporange}{audio encoders}, \textcolor{mpgreen}{music encoders}, \textcolor{mpbrown}{LALM audio encoders}, \textcolor{mpgray}{time series encoders}, \textcolor{mppurple}{signal encoders} and \textcolor{mpred}{our model} respectively.
\end{tablenotes}
\end{threeparttable}
}
\end{table*}
\endgroup

\subsection{Anomaly Detection}
\label{subsec:anomaly}

Anomaly detection is to predict whether a signal is normal or anomalous when no anomalies are provided for training, which emphasizes the scarcity of fault data. The model is evaluated on the datasets of the annual DCASE ASD challenge, including DCASE20~\cite{Koizumi_WASPAA2019_01, Purohit_DCASE2019_01, Koizumi_DCASE2020_01}, DCASE21~\cite{Tanabe_WASPAA2021_01, Kawaguchi2021, Harada2021}, DCASE22~\cite{Dohi2022, Dohi2022-2}, DCASE23~\cite{Dohi_arXiv2023_01, Harada_arXiv2023_01}, DCASE24~\cite{Nishida_arXiv2024_01, Harada_EUSIPCO2023_01} and DCASE25~\cite{Nishida_arXiv2025_01}. We use the official split and evaluate the model by the challenge criteria, which is based on AUC. For each dataset, we report the harmonic mean over both the development and the evaluation subsets. All models adopt the KNN-based anomaly detection pipeline in AnoPatch~\cite{jiang2024anopatch}.

\subsection{Fault Diagnosis}
\label{subsec:fault}

Fault diagnosis is to identify the specific fault type or health state of a signal with labeled data provided in advance. The fault diagnosis part is sourced from seven publicly available datasets: IDMT-ISA-COMPRESSED-AIR (IICA)~\cite{johnson2020compressed}, IDMT-ISA-ELECTRIC-ENGINE (IIEE)~\cite{grollmisch2019sounding}, the WT-planetary-gearbox-dataset (WTPG)~\cite{liu2023review}, the Machinery Fault Dataset (MaFaulDa)~\cite{ribeiro2016mafaulda}, the UM-GearEccDataset (UMGED)~\cite{li2025comprehensive}, the SDUST dataset~\cite{WANG2024111285, ZHANG2024102730, 10146304, HAN2022111131} and the PU Dataset~\cite{lessmeier2016condition}. To reveal the modality-specific performance, we first divide these data by the modality, resulting in 13 datasets. For each dataset, we flatten all multi-channel signals into single-channel and split them into segments if they are longer than 10~s. For KNN inference, $k$ is set to 5 for all models. All datasets are evaluated by macro-average accuracy.

To ensure proper difficulty of the task, we conduct sealed train-test split for datasets without official train-test splits, where segments from the same channel of the same recording cannot be partitioned into both the training and the test sets. Industrial signals are sometimes extremely stationary along the time axis due to the constant working conditions, causing segments to be highly identical. If random splitting were adopted, these segments would appear in both the training and test sets, causing the task to be exceedingly simple (99\%+) and making it hard to compare the true capabilities. This problem is also noted in~\cite{spadini2024intelligent}. Thus, we trace each channel of the original recording and allocate all its segments into either the training set or the test set\footnote{For the PU dataset, the signal under each working condition is recorded 20 times, resulting in 20 highly identical segments. Thus, we allocate these segments based on the working condition.}. The train-test split ratio is default to 1:1, except for UMGED datasets which are 4:1. We ensure that all classes are presented in both sets and evaluate the model under 10 different splits. The split ratio is further analyzed in Section~\ref{subsubsec:multi_split} to eliminate its interference.

\section{Experiment}
\label{sec:exp}

\subsection{Implementation Details of FISHER}
\label{subsec:train_detail}

\begin{table}[t]
\resizebox{\linewidth}{!}{
\centering
\begin{threeparttable}
\caption{Unique Hyperparameters of FISHER}
\label{tab:hyperparam}
\begin{tabular}{*{9}c}
\toprule
    \multirow{2}*{Variant} & Num & \multirow{2}*{$f_{base}$} & \multirow{2}*{$w$} & \multirow{2}*{$d$} & Num & Batch & \multirow{2}*{$m_b$\tnote{1}} & \multirow{2}*{$m_l$} \\
    & Param & & & & Head & Size\tnote{1} & & \\
    \midrule
    tiny & 5.5M & 4000 & 100 & 192 & 3 & 24 & 64 & 5 \\
    mini & 10M & 4000 & 100 & 256 & 4 & 32 & 64 & 5 \\
    small & 22M & 2000 & 50 & 384 & 6 & 16 & 128 & 2 \\
    \bottomrule
\end{tabular}
\begin{tablenotes}
\item[1] Both the batch size and $m_b$ are specified for each GPU.
\end{tablenotes}
\end{threeparttable}}
\end{table}

FISHER is trained under three scales, namely tiny (5.5M), mini (10M) and small (22M), which are in line with the hierarchy of ViT. Table~\ref{tab:hyperparam} lists unique hyperparameters for each version. As for the shared hyperparameters, $t_{win}$ and $t_{hop}$ are fixed as 25~ms and 10~ms, $f_{max}$ is 32~kHz, the mask ratio is 80\%, and $\lambda$ is 0.5. EMA decay $\tau$ is 0.9999. $E_{stu}$ contains 12 layers, and the patch size is $16{\times}16$. $D_{stu}$ is a 5-layer 2D convolutional network. All models are pre-trained on the combined dataset of Audioset~\cite{gemmeke2017audio}, Freesound\footnote{\url{https://freesound.org/}}, MTG-Jamendo~\cite{bogdanov2019mtg} and Music4all~\cite{santana2020music4all} with a total volume of 17k hours, which is indeed a dramatic breakthrough compared with previous works. We train each model for 400k steps on four NVIDIA RTX A6000 GPUs. For each model, we adopt a warm-up scheduler with a peak learning rate of 0.001 and a warm-up step of 40k. We use the sinc\_interp\_hann method in torchaudio to do random resampling in pre-training.

\subsection{Baselines}
\label{subsec:baselines}

We are curious about 1) how well SOTA encoders can encode industrial signals 2) training on what kinds of data are effective. We select 24 baselines, including two speech encoders~\cite{baevski2020wav2vec,radford2023whisper}, seven audio encoders~\cite{koutini2022efficient,huang2022masked,pmlr-v202-chen23ag,bharadwaj2025openbeats,chen2024eat,dinkel2023ced,dinkel2024scaling}, two music encoders~\cite{li2024mert,zhu2025muq}, four LALM audio encoders~\cite{chu2024qwen2,xu2025qwen2.5omni,ghosh2026audio,dinkel2025midashenglm}, two time series encoders~\cite{shi2025time,liu2025sundial}, and seven signal encoders~\cite{wu2009ensemble,bandt2002permutation,chen2024tfpred,yan2024liconvformer,peng2025bearllm,peng2025unified,chen2026cows,zhang2026echo}. Details are in Appendix~\ref{appen_sec:baseline}.

\subsection{Results on the RMIS Benchmark}
\label{subsec:result_RMIS}

Figure~\ref{fig:RMIS_all} compares the scaling curves on RMIS, while Table~\ref{tab:RMIS_num} presents the numerical results. The higher the score is, the better the model is. FISHER is the most versatile model for industrial signal analysis, achieving the highest RMIS score of 62.23\% and surpassing all baselines by at least 1.56\%. The second-best model, ECHO~\cite{zhang2026echo}, is a follow-up work of FISHER with the identical sub-band modeling scheme. After excluding it, the performance gain can further reach 4.2\%, suggesting that the superiority of FISHER has also been verified in other works. Moreover, FISHER is especially skilled in fault diagnosis tasks, where signals are mostly~recorded at high sampling rates. The performance gain of FISHER is contributed by the ability to seamlessly adapt to variable sampling rates, while baselines lose critical information during resampling. Finally, FISHER achieves superior scores with much smaller model sizes (up to 16x compared to common 90M encoders), which is better suited for real-world deployment. Aside from FISHER, specialized signal encoders (RotLLM, CoWS, etc) lag substantially behind out-of-domain pre-trained models. This suggests that large-scale pre-training can tackle heterogeneity even more severe than the M5 problem. Meanwhile, models with longer context windows (10 s) generally outperform those with only a few thousand points, revealing the importance of long-range dependencies for signal modeling.

\subsection{Extended Validation on RMIS}

\subsubsection{Multiple Split Ratios}
\label{subsubsec:multi_split}

\begin{figure}
\centering
\subfloat[Split Ratios]{
    \includegraphics[width=0.47\linewidth]{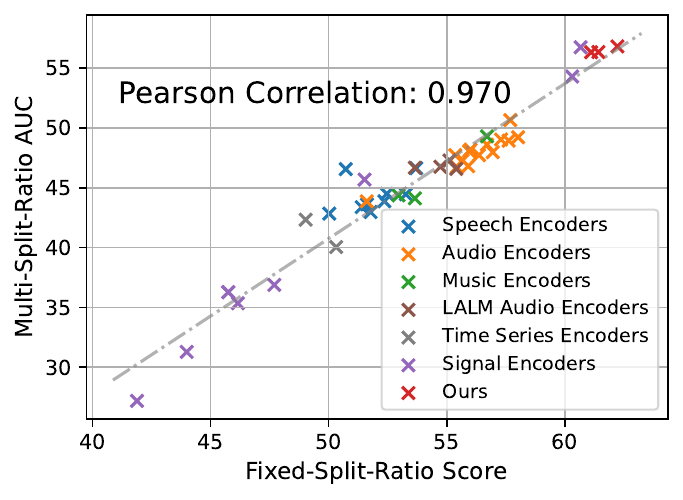}
    \label{fig:fix_ms_corr}
}
\hfill
\subfloat[Data Sources]{
    \includegraphics[width=0.47\linewidth]{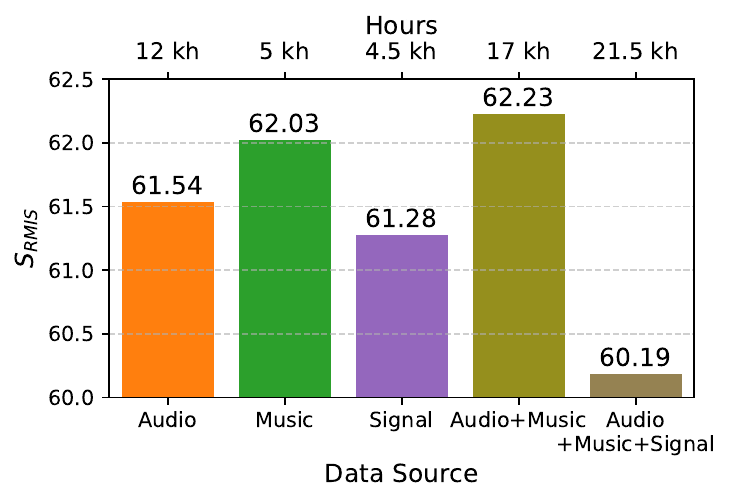}
    \label{fig:data_select}
}
\caption{Experiments on Split Ratios and Data Sources}
\label{fig:exp_comb}
\end{figure}

In the RMIS benchmark, 12 out of 13 fault diagnosis datasets do not provide official split. Previous works on these datasets adopt various split ratios. To eliminate the interference caused by the split ratio, we first evaluate each model under ascending split ratios from 0.05 to 0.95 (train set ratio), where for each ratio, the experiment is conducted under 10 different splits. Then for each dataset, we plot the performance curve with respect to the split ratio and estimate the area under it. Finally, these AUCs are averaged across datasets. Figure~\ref{fig:fix_ms_corr} plots the fixed-split-ratio score and the multi-split-ratio AUC for each model, which reveals a strong correlation between the two split settings. Thus, it is reasonable to evaluate the model only under a fixed split ratio.

\subsubsection{Few-Shot KNN}

\begin{table}[t]
\centering
\caption{Few-shot Fault Diagnosis Performances}
\label{tab:few_shot}
\resizebox{\linewidth}{!}{
\begin{tabular}{*{8}{c}}
\toprule
Model & Variant & 1 Shot & 2 Shots & 3 Shots & 4 Shots & 5 Shots & Mean \\
\midrule
\textcolor{mporange}{\textbf{BEATs}} & iter3 & 25.04 & 27.58 & 29.43 & 30.83 & 32.04 & 28.98 \\
\spacedashline
\multirow{2}*{\textcolor{mporange}{\textbf{EAT}}} & base30 & 26.82 & 29.49 & 31.41 & 32.90 & 34.12 & 30.95 \\
& large10 & 27.65 & 30.20 & 31.98 & 33.41 & 34.58 & 31.56 \\
\spacedashline
\multirow{4}*{\textcolor{mporange}{\textbf{CED}}} & tiny & 27.96 & 30.37 & 32.05 & 33.38 & 34.45 & 31.64 \\
& mini & \bstscr{29.26} & 31.76 & 33.44 & 34.76 & 35.83 & 33.01 \\
& small & 29.16 & 31.70 & 33.41 & 34.72 & 35.81 & 32.96 \\
& base & 29.20 & 31.89 & 33.70 & 35.11 & 36.24 & 33.23 \\
\spacedashline
\multirow{3}*{\textcolor{mporange}{\textbf{DaSheng}}} & base & 26.71 & 29.15 & 30.91 & 32.32 & 33.48 & 30.51 \\
& 0.6B & 26.59 & 28.84 & 30.50 & 31.85 & 32.95 & 30.15 \\
& 1.2B & 26.70 & 28.86 & 30.48 & 31.80 & 32.83 & 30.13 \\
\spacedashline
\multirow{3}*{\textcolor{mpred}{\textbf{FISHER}}} & tiny & 27.67 & 30.99 & 33.66 & 35.92 & 37.75 & 33.20 \\
& mini & 27.97 & 31.32 & 34.00 & 36.24 & 38.09 & 33.52 \\
& small & 28.50 & \bstscr{32.05} & \bstscr{34.85} & \bstscr{37.14} & \bstscr{38.98} & \bstscr{34.30} \\
\bottomrule
\end{tabular}}
\end{table}

Apart from anomaly detection (no fault as reference) and 1:1 fault diagnosis (abundant fault as reference), fault diagnosis with few fault data is also vital. To this end, we reorganize RMIS fault diagnosis datasets by a few-shot setting~\cite{wang2019simpleshot}. For each dataset, we randomly sample $K\in\{1,2,3,4,5\}$ and $Q=15$ segments for each class to form the support set and the query set respectively, where sealed train-test split is enforced to prevent information leakage. KNN ($k=1$) is employed to classify the query set based on the support set, and the model is evaluated for 1500 random splits (episodes). Table~\ref{tab:few_shot} compares the few-shot fault diagnosis performances of top models, where FISHER outperforms these baselines by at least 0.61\%, showcasing superior scalability across no fault, few fault and abundant fault scenarios.

\begin{figure}[t]
\centering
\subfloat[FISHER-tiny]{
    \includegraphics[width=0.305\linewidth]{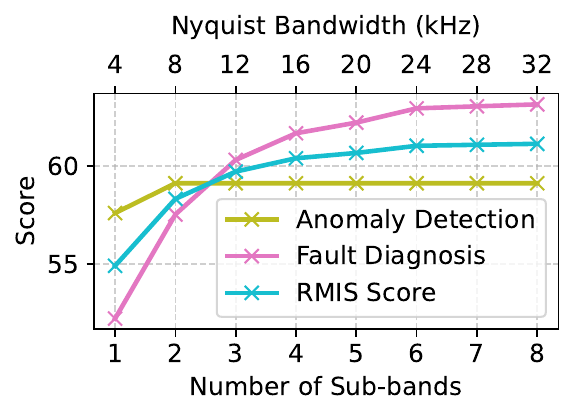}
}
\hfill
\subfloat[FISHER-mini]{
    \includegraphics[width=0.305\linewidth]{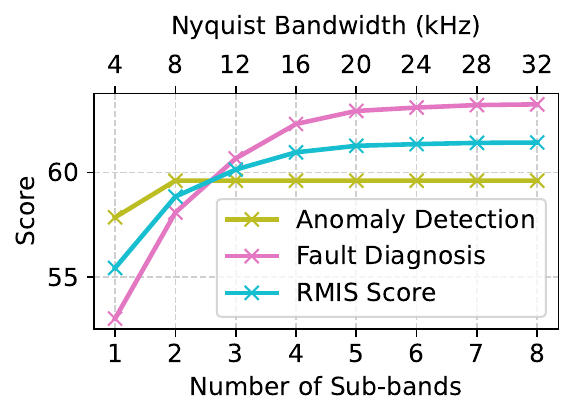}
}
\hfill
\subfloat[FISHER-small]{
    \includegraphics[width=0.305\linewidth]{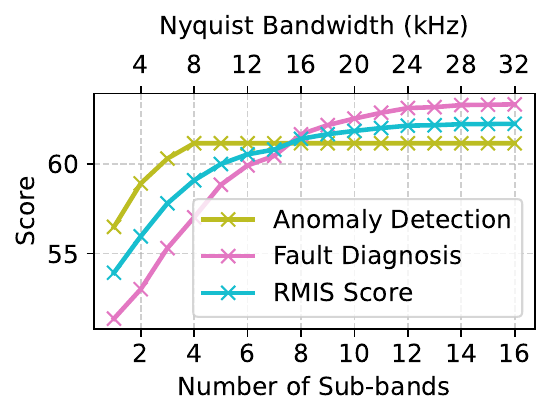}
}
\caption{Performance of FISHER vs. Number of Sub-bands}
\label{fig:fisher_score_nb}
\end{figure}

\subsection{Key Findings}

\subsubsection{High Frequency Gain}
\label{subsec:high_freq_gain}

PaSST~\cite{koutini2022efficient}, ECHO~\cite{zhang2026echo} (follow-up work of FISHER), and FISHER are the only large-scale pre-trained models that accept sampling rates no less than 32 kHz. These models conformably surpass respective contemporary models on RMIS, suggesting that high-frequency components are crucial for analyzing industrial signals. To validate it, we constrain the number of sub-bands available to FISHER (starting from low frequency) and evaluate it on RMIS. As depicted in Fig.~\ref{fig:fisher_score_nb}, the score grows steadily until it reaches the sampling rate of the anomaly detection datasets (16~kHz). On fault diagnosis datasets with higher sampling rates, the score continues to grow monotonically. This suggests that the success of FISHER is mainly attributed to its ability to utilize full bandwidths of multiple sampling rates, and this ability is crucial for developing signal foundation models.

\begin{figure}[t]
\centering
\subfloat[Audio]{
    \includegraphics[width=0.3\linewidth]{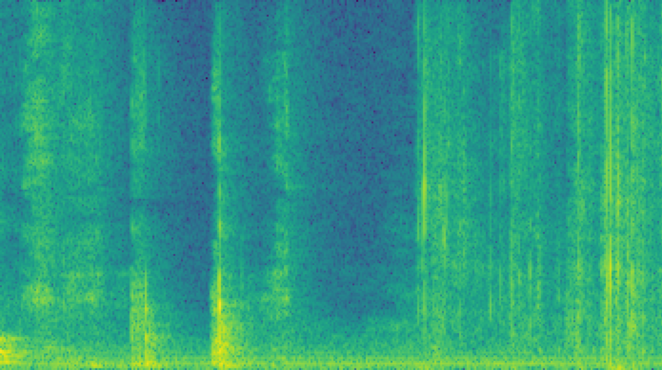}
}
\subfloat[Music]{
    \includegraphics[width=0.3\linewidth]{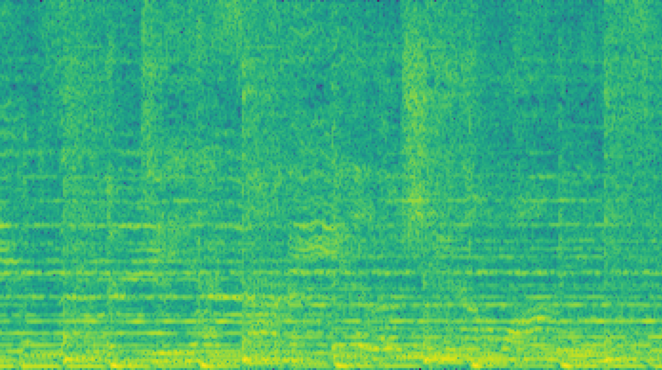}
}
\subfloat[Industrial Signals]{
    \includegraphics[width=0.3\linewidth]{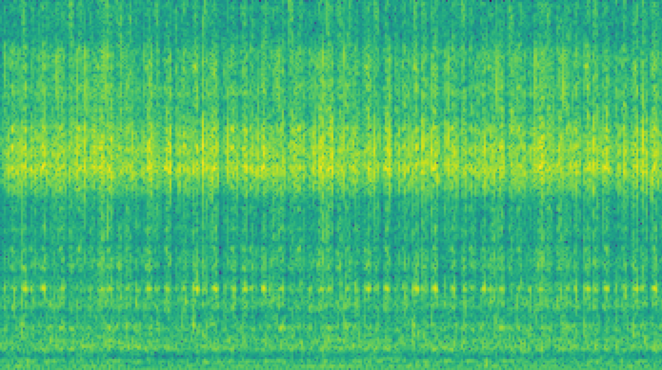}
}
\caption{STFT of Different Data Sources}
\label{fig:stft_sour}
\end{figure}

\subsubsection{Data Curation for Pre-training}
\label{subsec:data_select}

The successful applications of audio and music pre-trained models on the RMIS benchmark indicate that signal models could benefit from training on audio and music data. In fact, as illustrated in Fig.~\ref{fig:stft_sour}, audio and music are more variant and random in temporal, which are of better quality for pre-training than stationary industrial signals. To validate it, we train FISHER-small on different data sources, where audio data are sourced from Audioset and Freesound, while music data are sourced from MTG-Jamendo and Music4all. Due to the limited amount of open-sourced industrial signal data, we self-collect sound and vibration signals from five types of real machines: coal grinder, pump, transformer, water turbine, and wind turbine. Then, a CED-tiny model is employed for data deduplication, resulting in a filtered dataset of 4.5k hours.

Figure~\ref{fig:data_select} compares the efficacy of these data sources. For a single source, music is better than audio, which in turn is better than industrial signals. For combined sources, combining audio and music yields the highest score of 62.23\%, whereas further adding industrial signals leads to a notable decrease. These results suggest that an industrial signal foundation model does not necessarily need to be trained on industrial signals. Audio and music data are easier to obtain and more effective for pre-training, and thus are more preferred currently. However, the model pre-trained on pure industrial signals does not lag far behind. Thus, how to align multi-modal series data and unleash joint efforts will be a key point, where more thorough data cleaning is required.

\begin{figure}[t]
\centering
\subfloat[Spectrogram and Band Splitting]{
    \includegraphics[width=0.47\linewidth]{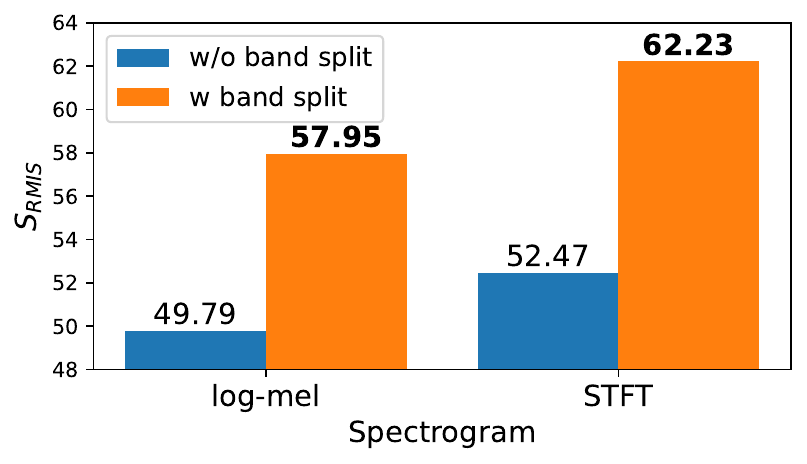}
    \label{fig:abla_spec_band}
}
\hfill
\subfloat[Bandwidth $w$]{
    \includegraphics[width=0.47\linewidth]{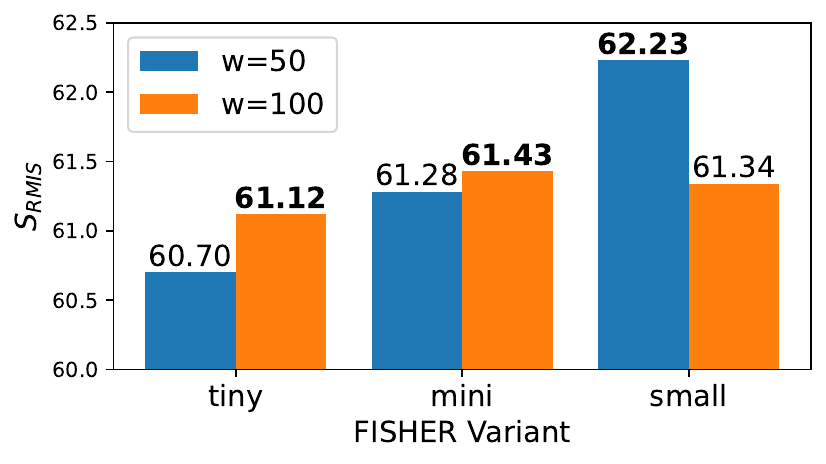}
    \label{fig:abla_bandwidth}
}

\subfloat[Pre-training Sampling Rate]{
    \includegraphics[width=0.47\linewidth]{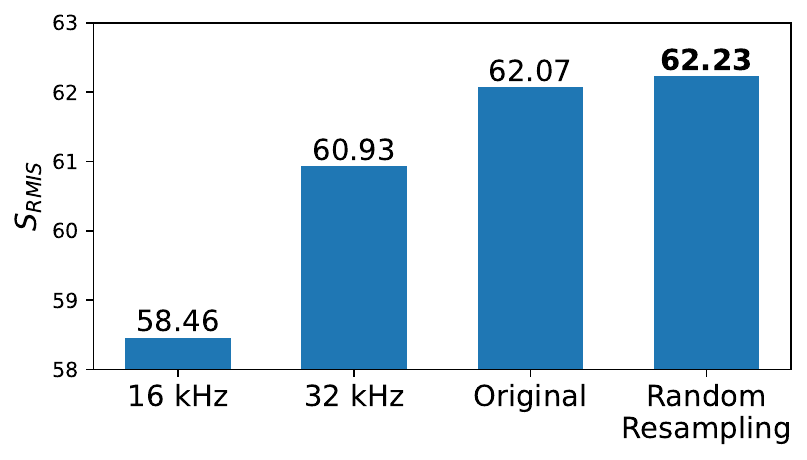}
    \label{fig:abla_train_sr}
}
\hfill
\subfloat[STFT $t_{win}$ and $t_{hop}$]{
    \includegraphics[width=0.47\linewidth]{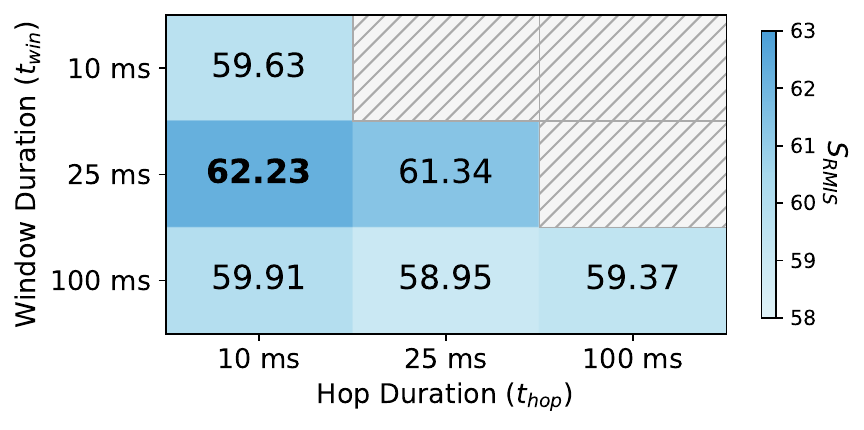}
    \label{fig:abla_spec_params}
}

\subfloat[Loss Combination $\lambda$]{
    \includegraphics[width=0.47\linewidth]{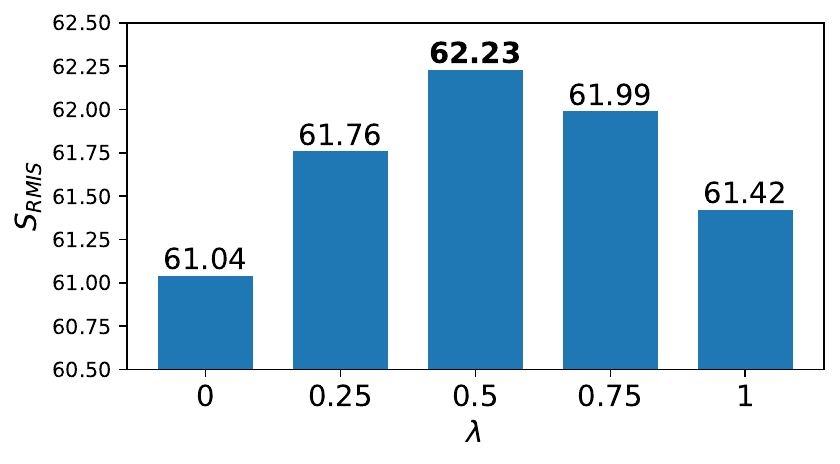}
    \label{fig:abla_loss_comb}
}
\hfill
\subfloat[Mask Cloning $m_b$]{
    \includegraphics[width=0.47\linewidth]{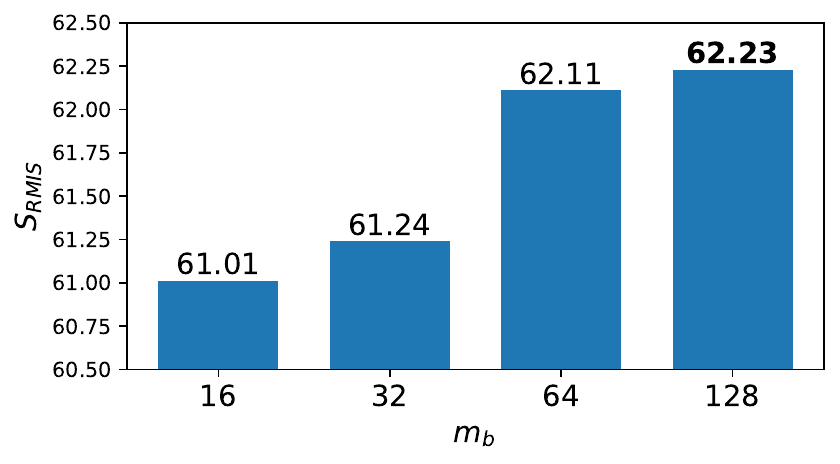}
    \label{fig:abla_mask_clone}
}
\caption{Ablation Study}
\label{fig:abla}
\end{figure}

\subsection{Ablation Study}

\subsubsection{Spectrogram and Band Splitting}

Figure~\ref{fig:abla_spec_band} validates the effectiveness of STFT and band splitting. Band splitting yields huge improvements of 8.16\% and 9.76\% for both spectrograms, which is surely the most effective design. Adopting STFT also yields an improvement of 4.28\%, which aligns perfectly with the theoretical analysis in Section~\ref{subsec:sb_model}.

\subsubsection{Bandwidth}

Figure~\ref{fig:abla_bandwidth} experiments with different bandwidth $w$ for three FISHER variants, where all scores are considerably higher than baselines, suggesting that the band-splitting scheme is robust to $w$. It is worthy to note that the tiny and mini variants favor larger bandwidths, whereas the small variant performs better with a smaller one.

\subsubsection{STFT Window Size and Hop Size}

Figure~\ref{fig:abla_spec_params} experiments with different $t_{win}$ and $t_{hop}$ on FISHER-small, where the optimal setting is (25 ms, 10 ms). This is likely due to the inductive bias of the pre-training data (audio \& music): The optimal is achieved when the STFT resolution aligns with the inherent characteristic of the pre-training data.

\subsubsection{Pre-training Sampling Rate}

Figure~\ref{fig:abla_train_sr} compares different pre-training sampling rates for FISHER-small, where $S_{RMIS}$ grows drastically and monotonically with the data sampling rate. This is consistent with the conclusion of Section~\ref{subsec:high_freq_gain}. Meanwhile, random resampling in pre-training further improves the scalability across different sampling rates.

\subsubsection{Loss Combination}

FISHER is self-distilled from two levels. Figure~\ref{fig:abla_loss_comb} experiments with different $\lambda$ on FISHER-small. The best score is achieved at $\lambda=0.5$, suggesting that $L_{band}$ and $L_{patch}$ are equally important.

\subsubsection{Mask Cloning}

Mask cloning enlarges the training data by creating different views of a sub-band. Figure~\ref{fig:abla_mask_clone} experiments with different $m_b$ on FISHER-small, where $S_{RMIS}$ grows monotonically with $m_b$. Thus, we set $m_b$ to the maximum value allowed by video random access memory (VRAM).

\subsubsection{SSL Framework}

Aside from the self-distillation framework ($S_{RMIS}=62.23$), we also pre-train the band splitting scheme with the AudioMAE framework, resulting in $S_{RMIS}=57.56$ ($51.61$ for vanilla AudioMAE). This suggests that 1) the self-distillation framework is better for SSL training 2) the band splitting scheme is robust across SSL frameworks.

\subsection{Inference Efficiency}

\begin{figure}
\centering
\includegraphics[width=0.945\linewidth]{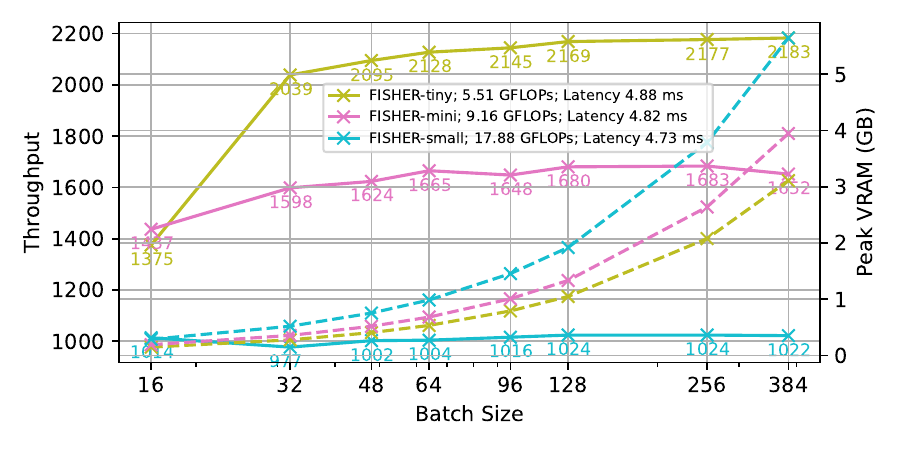}
\caption{Inference Efficiency of FISHER on a 3090 GPU. Solid lines denote throughput, and dashed lines denote peak VRAM.}
\label{fig:inf_eff}
\end{figure}

FISHER models the sub-band instead of the whole spectrogram, which reduces the need for VRAM, allowing us to deploy the model on a consumer-grade GPU. To quantify this, four metrics are evaluated: Giga floating point operations (GFLOPs), latency (p50), throughput, and peak VRAM. FISHER is evaluated on 16 kHz signal segments (10~s) with FP16 inference. Since its complexity grows linearly with the number of sub-bands, results on other sampling rates can be scaled linearly. The experiment is conducted on an RTX 3090 GPU. Figure~\ref{fig:inf_eff} presents the efficiency curves of FISHER, suggesting that FISHER can process signals from thousands of sensors per second on an entry-level GPU, and such overhead is negligible compared to the malfunctional losses it can save.

\section{Conclusion}
\label{sec:conclu}
In this paper, we first formulated the M5 problem and derived a prioritized solution roadmap. We then proposed FISHER, which models the information gain of higher sampling rates as the concatenation of sub-band information. We also developed the RMIS benchmark to evaluate out-of-the-box diagnostic abilities, where FISHER exceeds all baselines by a wide margin with much smaller sizes. We further drew two key findings on high frequency gain and data curation.

\section*{Acknowledgment}

We thank Cheng Lu and Yihong Qiu for their suggestions.

\bibliographystyle{IEEEtran}
\bibliography{refer.bib}

\appendices


\begingroup
\setlength{\tabcolsep}{3.5pt}  
\begin{table}[b]
\centering
\caption{Detailed Task Setup for Fault Diagnosis Datasets}
\label{tab:fd_setup}
\adjustbox{max width=\linewidth}{
\begin{tabular}{ccl}
\toprule
Dataset & Task & Class Setup \\
\midrule
\multirow{2}*{IICA} & Air Compressor & tubeleak\_iO, tubeleak\_niO, ventleak\_iO,\\
& Leakage & ventleak\_niO, ventlow\_iO, ventlow\_niO\\
\midrule
IIEE & Electric Engine Fault & good, broken, heavy load\\
\midrule
\multirow{2}*{WTPG} & Planetary & broken, healthy, missing tooth,\\
& Gearbox Fault & root crack, wear \\
\midrule
\multirow{3}*{MaFaulDa} & \multirow{3}*{\makecell[c]{Bearing\\Fault}} & normal, horizontal misalignment, vertical misalignment, imbalance,\\
& & underhang cage fault, underhang outer race, underhang ball fault,\\
& & overhang cage fault, overhang outer race, overhang ball fault\\
\midrule
SDUST & Bearing & NC, OF0.2, OF0.4, OF0.6, IF0.2\\
bearing & Fault & IF0.4, IF0.6, RF0.2, RF0.4, RF0.6\\
\midrule
SDUST & Planetary & NC, planetary fracture, planetary pitting,\\
gear & Gearbox Fault & planetary wear, sun fracture, sun pitting, sun wear\\
\midrule
UMGED & Gear Eccentricity & E00, E02, E04, E06, E08, E10, E12, E14, E16, E18, E20\\
\midrule
PU & Bearing Fault & healthy, IR, OR\\
\bottomrule
\end{tabular}}
\end{table}
\endgroup

\section{Details of the RMIS Benchmark}
\label{appsec:rmis_detail}

For ASD tasks, all models adopt the identical KNN-based detection pipeline as AnoPatch~\cite{jiang2024anopatch} after feature extraction, where normal embeddings from the training set form memory banks, and the anomaly score of each query embedding of the test set is calculated as the average distance to the nearest neighbors. The distance metric is selected as cosine distance and $k$ is kept as 1. To reveal the intrinsic ability of the model, we do not tune the hyperparameters of KNN on each ASD dataset. For the DCASE20 dataset, anomaly detection is conducted per machine id. For the rest DCASE datasets, anomaly detection is conducted per section. Since domain shift is involved, two memory banks are constructed for each section, one for the source embeddings and the other for the target embeddings.
For fault diagnosis tasks, Table~\ref{tab:fd_setup} presents the detailed task setup and class mapping.

\section{Baseline Details}
\label{appen_sec:baseline}

For each model, we follow the official embedding procedure and select the top performing open-sourced checkpoints. Embeddings are extracted from the last layer. If the model has a special token for representing the overall signal, e.g. [CLS] token, it is utilized for evaluation. If not, frame-level embeddings are mean-pooled to obtain the utterance-level embeddings. Detailed checkpoint selections are posted below for models with multiple checkpoints.

\subsubsection{Speech Encoders}
For Wav2Vec 2.0, we employ the official wav2vec2-base-960h, wav2vec2-xls-r-300m, wav2vec2-xls-r-1b, wav2vec2-x1s-r-2b checkpoints. For Whisper, we evaluate the encoders of five official pre-trained checkpoints.

\subsubsection{Audio Encoders}
For PaSST, we employ the official \underline{passt-s-f128-p16-s10-ap.476-swa} checkpoint. For OpenBEATs, we employ the base and large checkpoints of iter3.

\subsubsection{LALMs}
We evaluate the audio encoder of each LALM.

\subsubsection{Time Series Encoders}
Time series models are not designed to process kHz-level signals, since 1) they model raw time points, which is highly susceptible to noise 2) their input windows (thousands of points) are much too short to capture second-level dependency, and their computational complexity is unacceptable. Thus, only two models are compared.

\subsubsection{Signal Encoders}
For PEEMD, we follow the works of~\cite{wu2009ensemble,bandt2002permutation}, which combine IMFs extraction, instability detection and ensemble averaging. For LiConvFormer, we utilize the weights trained on CWRU. For BearLLM and RotLLM, we employ their feature encoders.

\vspace{-11.5ex}
\begin{IEEEbiography}[{\includegraphics[width=1in,height=1.25in,clip,keepaspectratio]{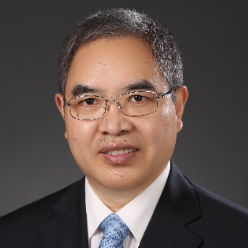}}]{Pingyi Fan}
(Senior Member, IEEE) is a tenured professor at the Department of Electronic Engineering of Tsinghua University, a US-NAAI member, an IEAS academician, and an IET fellow. His recent research covers 6G wireless communication networks and machine learning, semantic information theory and generalized information theory, big data processing theory, and intelligent network and system detection.
\end{IEEEbiography}
\vspace{-14ex}
\begin{IEEEbiography}[{\includegraphics[width=1in,height=1.25in,clip,keepaspectratio]{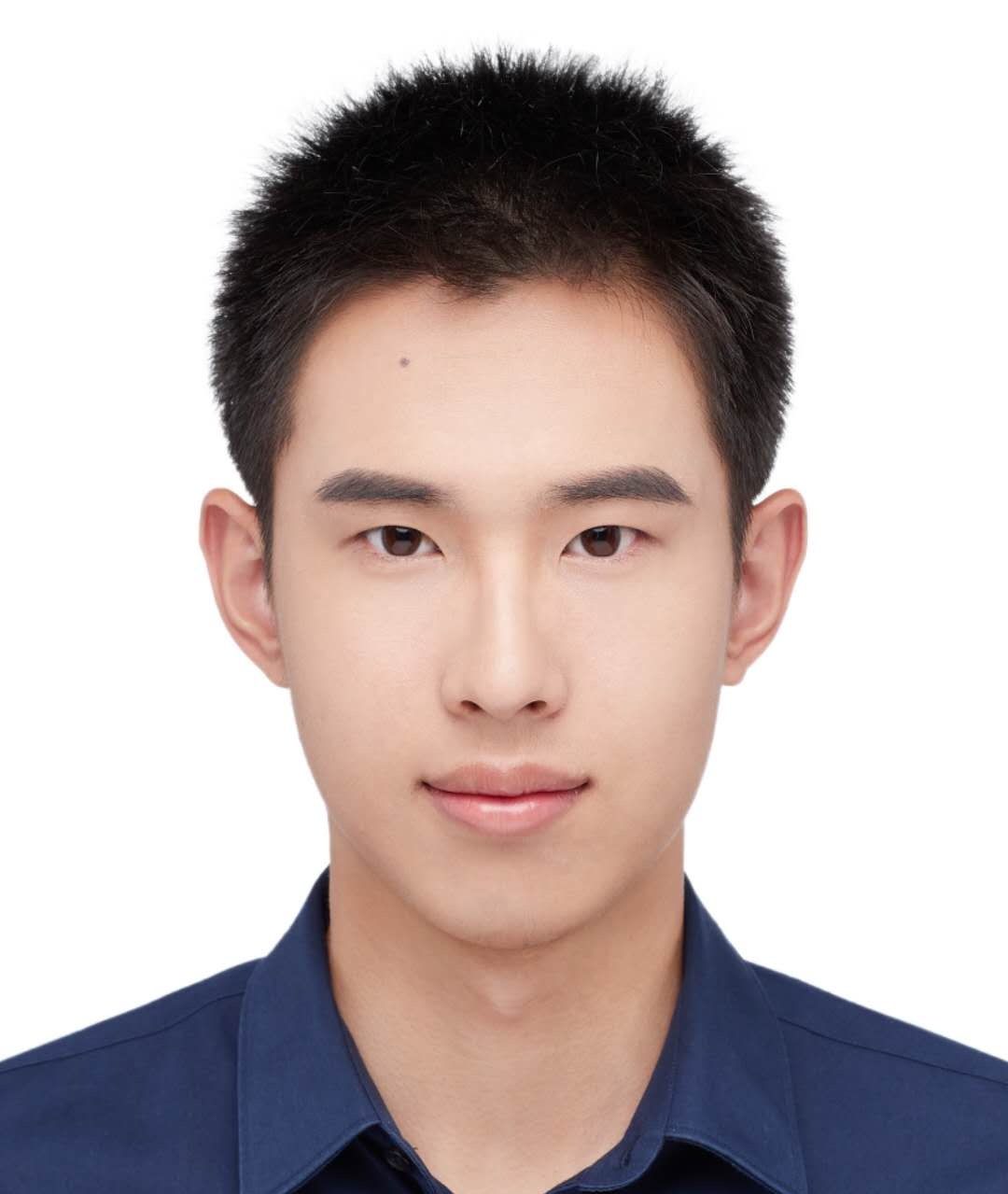}}]{Anbai Jiang}
received the B.Eng. degree in Electronic Engineering from Tsinghua University, Beijing, China, in 2022. He is currently a fourth-year Ph.D. student at the Department of Electronic Engineering of Tsinghua University, Beijing, China, under the supervision of Prof. Pingyi Fan. His current research focuses on industrial signal representation and analysis.
\end{IEEEbiography}
\vspace{-14ex}
\begin{IEEEbiography}[{\includegraphics[width=1in,height=1.25in,clip,keepaspectratio]{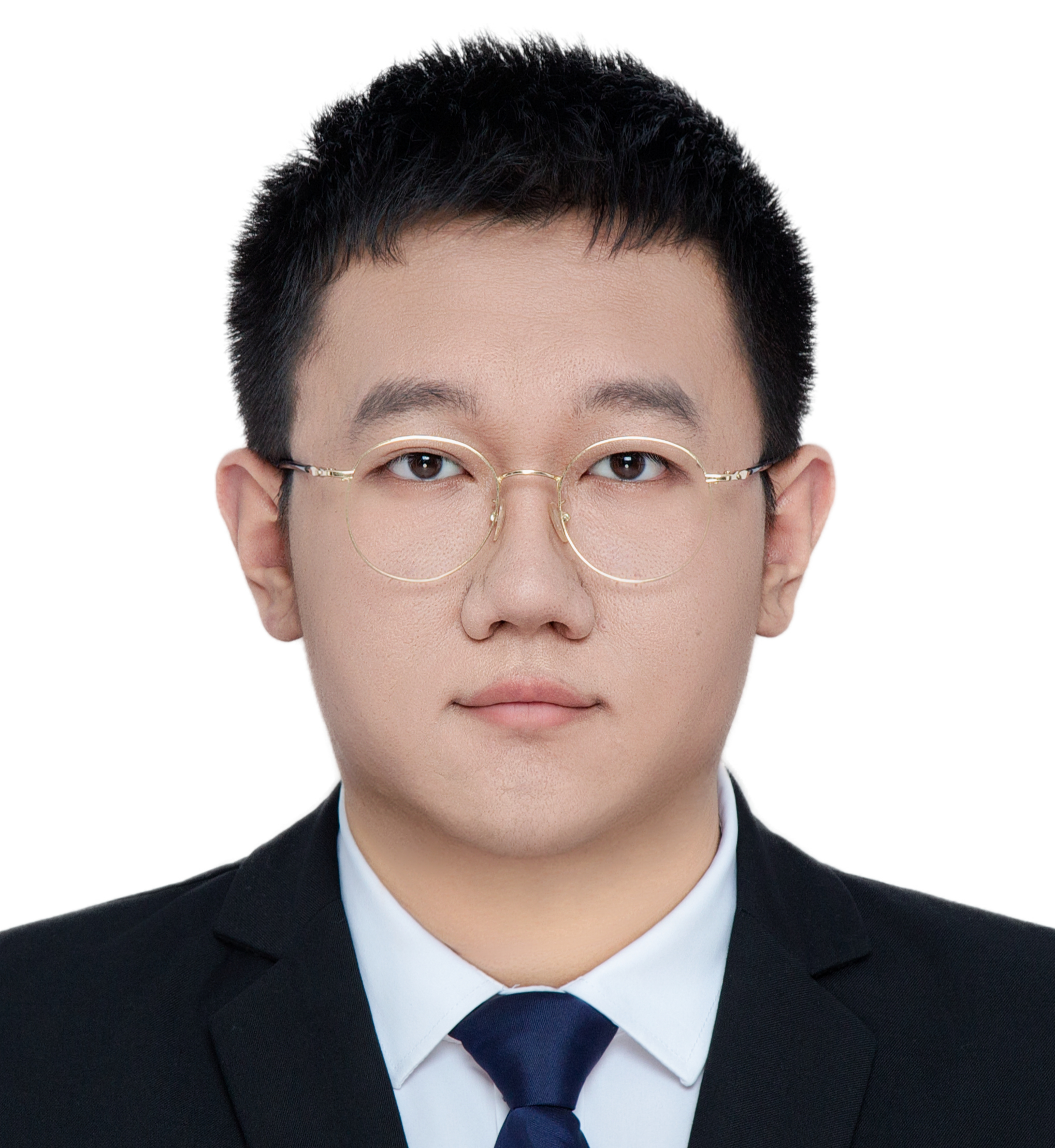}}]{Shuwei Zhang}
received the B.Eng. degree from the School of Information and Electronics at Beijing Institute of Technology, Beijing, China, in 2025. He is currently working toward the Ph.D. degree with the Department of Electronic Engineering, Tsinghua University, Beijing, China. His research interests include signal representation, analysis and generation.
\end{IEEEbiography}
\vspace{-14ex}
\begin{IEEEbiography}[{\includegraphics[width=1in,height=1.25in,clip,keepaspectratio]{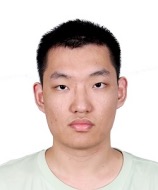}}]{Xinhu Zheng}
received the B.Eng. degree in Electronic Engineering from Tsinghua University, Beijing, China, in 2024. He is currently pursuing the Ph.D. degree in Computer Science at Shanghai Jiao Tong University, Shanghai, China, under the supervision of Prof. Yanmin Qian. His research focuses on deep learning-based sound anomaly detection and high-frequency signal analysis.
\end{IEEEbiography}
\vspace{-12ex}
\begin{IEEEbiography}[{\includegraphics[width=1in,height=1.25in,clip,keepaspectratio]{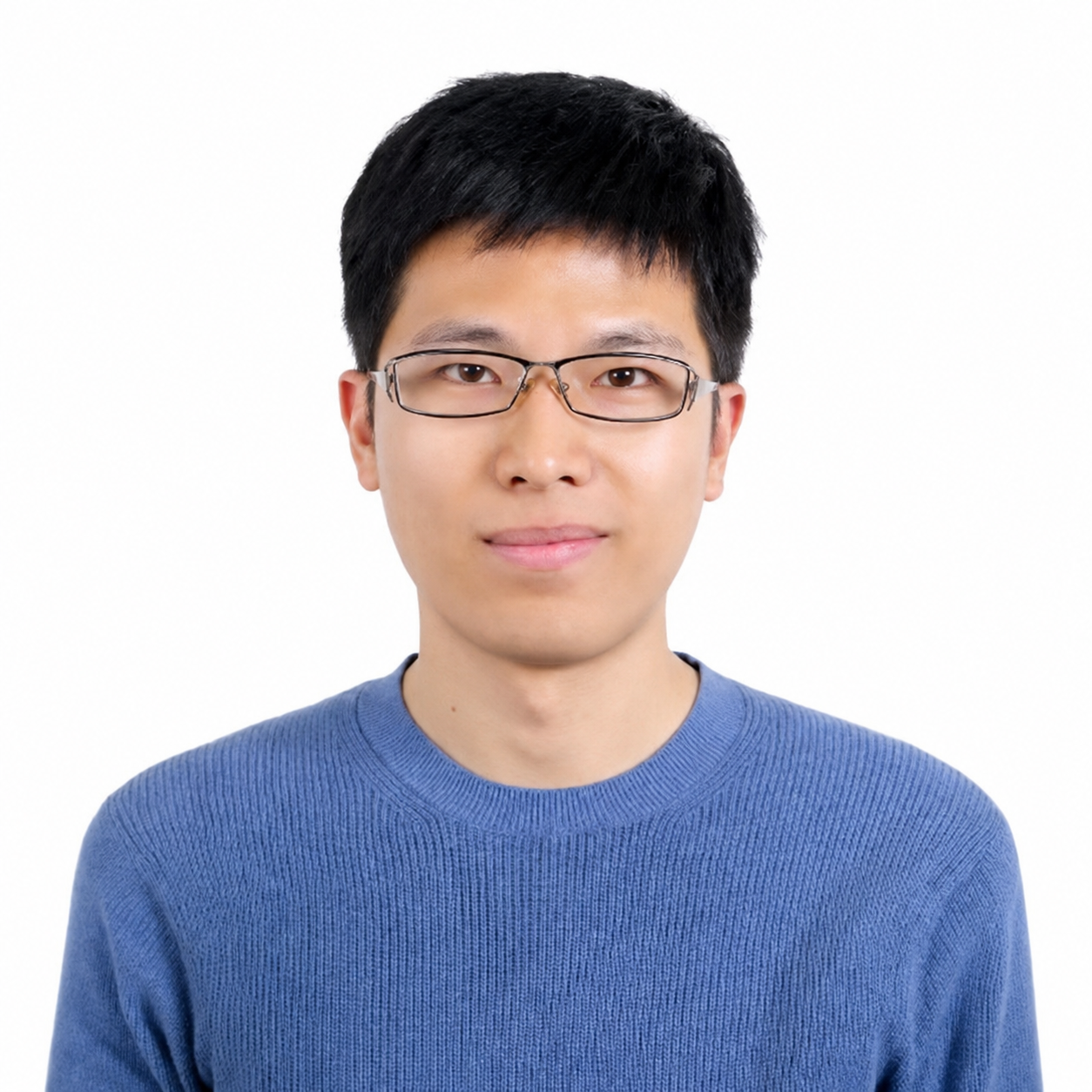}}]{Zhiqiang Lv}
(Member, IEEE) received the Ph.D. degree in Engineering from the Department of Electronic Engineering at Tsinghua University, Beijing, China, in 2018. His research focuses on speech understanding and spoken language interaction.
\end{IEEEbiography}
\vspace{-9.5ex}
\begin{IEEEbiography}[{\includegraphics[width=1in,height=1.25in,clip,keepaspectratio]{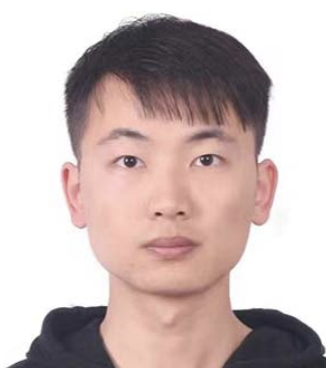}}]{Bing Han}
received the B.Eng.~degree from the Department of Computer Science and Engineering, Shanghai Jiao Tong University, Shanghai, China, in 2020. He is currently working toward the Ph.D.~degree in Shanghai Jiao Tong University, Shanghai, China, under the supervision of Prof. Yanmin Qian. His current research mainly focuses on speech and audio understanding.
\end{IEEEbiography}
\vspace{-10ex}
\begin{IEEEbiography}[{\includegraphics[width=1in,height=1.25in,clip,keepaspectratio]{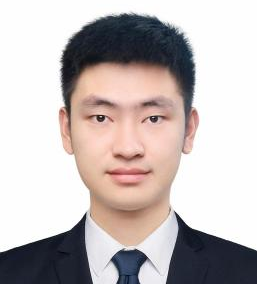}}]{Wenrui Liang}
(Student Member, IEEE) received the B.S. degree in Electronic Engineering from Shanghai Jiao Tong University, Shanghai, China, in 2024. He is currently working toward the M.S. degree with the Speech and Audio Technology Lab, Department of Electronic Engineering, Tsinghua University. His research interests include audio analysis and generation.
\end{IEEEbiography}
\vspace{-9.5ex}
\begin{IEEEbiography}[{\includegraphics[width=1in,height=1.25in,clip,keepaspectratio]{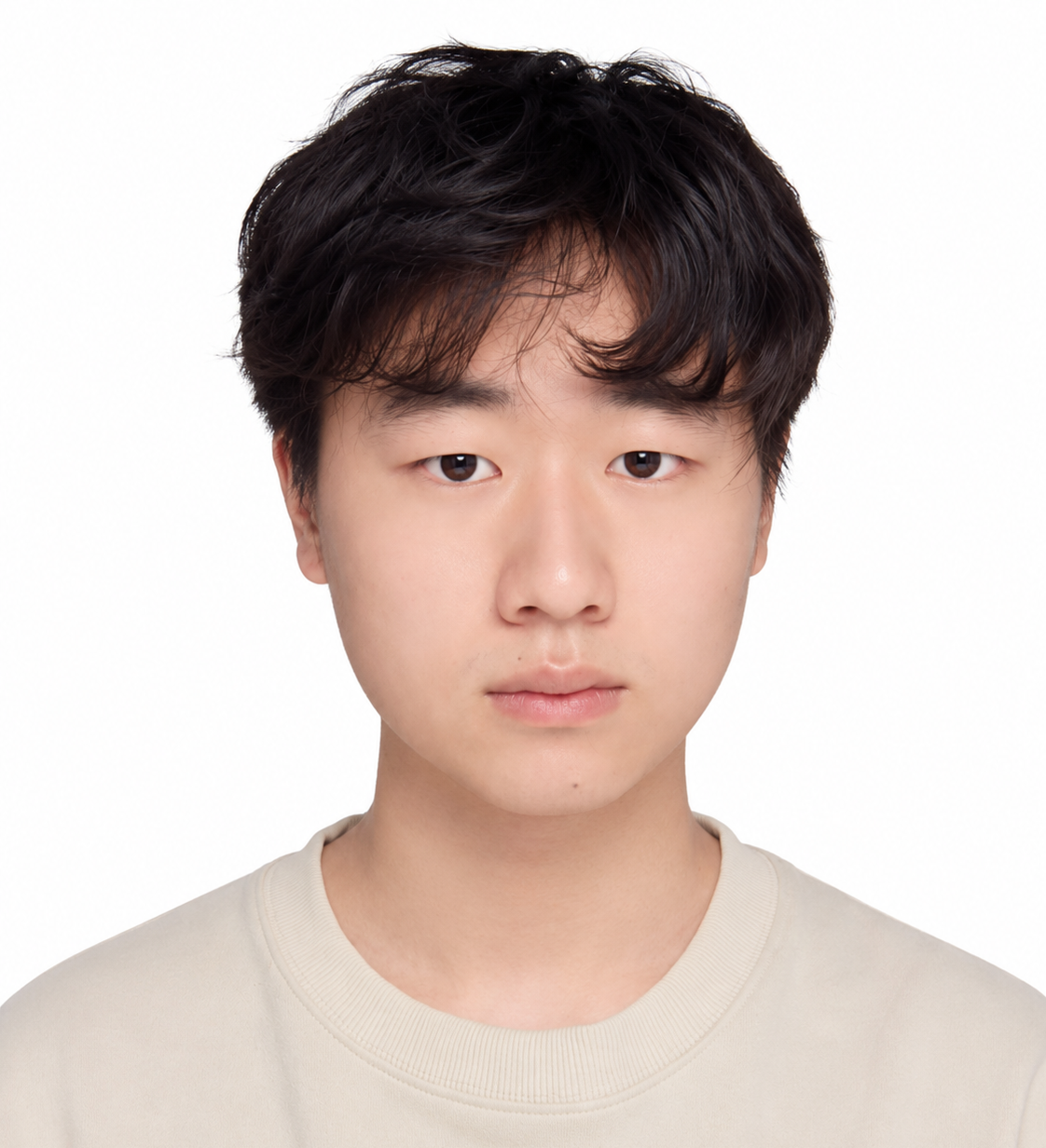}}]{Junjie Li}
is currently a senior undergraduate student in the Department of Electronic Engineering at Tsinghua University, Beijing, China. He will join Nanyang Technological University as a Ph.D. student in 2026. His current research interests include software security, logic vulnerability detection, container runtime security, and LLM-based vulnerability analysis and remediation.
\end{IEEEbiography}
\vspace{-10ex}
\begin{IEEEbiography}[{\includegraphics[width=1in,height=1.25in,clip,keepaspectratio]{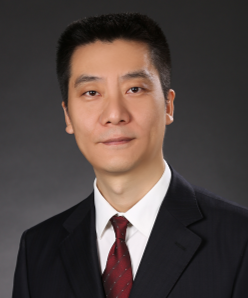}}]{Wei-Qiang Zhang}
(Senior Member, IEEE) is a Research Professor with the Department of Electronic Engineering, Tsinghua University, head of the Speech and Audio Technology Laboratory (SATLab), and director of the Research Center for Data and Computility, Institute for Embodied Intelligence and Robotics, Tsinghua University. He is a recipient of six ministerial and provincial-level Science and Technology or Teaching Achievement Awards.
\end{IEEEbiography}
\vspace{-10.2ex}
\begin{IEEEbiography}[{\includegraphics[width=1in,height=1.25in,clip,keepaspectratio]{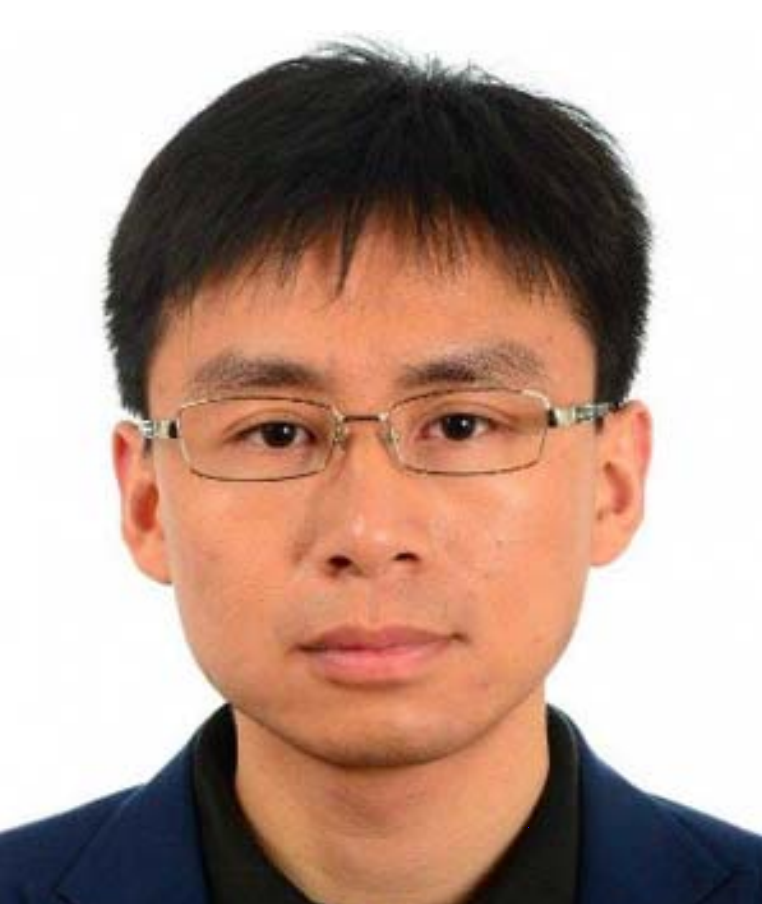}}]{Yanmin Qian}
(Senior Member, IEEE) is a full professor with the School of Computer Science, Shanghai Jiao Tong University, Shanghai, China. His research interests include speech and audio processing, automatic speech recognition and translation, speaker and language recognition, speech separation and enhancement, music generation and understanding, and multimodal information processing.
\end{IEEEbiography}
\vspace{-9.5ex}
\begin{IEEEbiography}[{\includegraphics[width=1in,height=1.25in,clip,keepaspectratio]{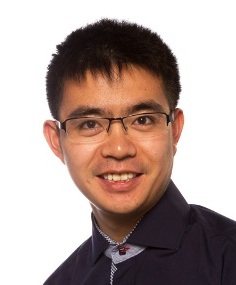}}]{Xie Chen}
(Senior Member, IEEE) is currently a Tenure-Track Associate Professor in the Department of Computer Science and Engineering at Shanghai Jiao Tong University, China. His main research interest lies in deep learning, especially its application to speech processing, including speech recognition and synthesis.
\end{IEEEbiography}
\vspace{-9.8ex}
\begin{IEEEbiography}[{\includegraphics[width=1in,height=1.25in,clip,keepaspectratio]{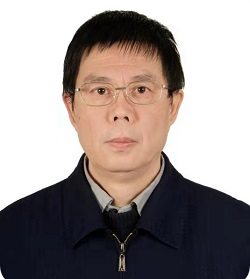}}]{Jia Liu}
(Member, IEEE) is a Professor in the Department of Electronic Engineering, Tsinghua University. He is also the Chief Scientist and the Co-Founder of HuaKong AI Plus Co., Ltd., Beijing, China. His research fields include speech recognition, speaker recognition, language recognition, expressive speech synthesis, speech coding, and spoken language understanding.
\vspace{-5ex}
\end{IEEEbiography}

\end{document}